\documentclass[runningheads]{llncs}
\usepackage{graphicx}

\usepackage{tikz}
\usepackage{comment}
\usepackage{amsmath,amssymb} %
\usepackage{color}

\usepackage[accsupp]{axessibility}  %

\usepackage{times}
\usepackage{soul}
\usepackage{url}
\usepackage[hidelinks]{hyperref}
\usepackage[utf8]{inputenc}
\usepackage[small]{caption}
\usepackage{graphicx}
\usepackage{booktabs}
\usepackage{algorithm}
\usepackage{algorithmic}
\usepackage{hyperref}
\usepackage{cleveref}
\usepackage{url}
\usepackage{multicol}
\usepackage{multirow}
\usepackage{adjustbox}
\usepackage{booktabs}
\usepackage{boldline}
\usepackage{mathtools}

\usepackage{epsfig}
\usepackage{subfiles}
\usepackage{color,soul}
\usepackage{caption}
\usepackage{subcaption}
\usepackage{array}
\usepackage[english]{babel}
\usepackage{comment}
\usepackage{multirow}
\usepackage{adjustbox}
\usepackage{boldline}
\newcolumntype{P}[1]{>{\centering\arraybackslash}p{#1}}
\newcolumntype{M}[1]{>{\centering\arraybackslash}m{#1}}
\usepackage{arydshln}

\newcommand{\eg}{\emph{e.g.,}}

\def\etal{\emph{et al.}}

\begin{document}
\pagestyle{headings}
\mainmatter
\def\ECCVSubNumber{5}  %

\title{Weakly Supervised Invariant Representation Learning Via Disentangling Known and Unknown Nuisance Factors} %

\titlerunning{Weakly invariant representation}
\author{Jiageng Zhu\inst{1,2,3} \and Hanchen Xie\inst{2,3} \and Wael Abd-Almageed\inst{1,2,3}}
\authorrunning{J. Zhu, H. Xie and W. Abd-Almageed}
\institute{USC Ming Hsieh Department of Electrical and Computer Engineering \and
USC Information Sciences Institute \and Visual Intelligence and Multimedia Analytics Laboratory\\
\email{\{jiagengz, hanchenx, wamageed\}@isi.edu}}
\maketitle

\newcommand{\fix}{\marginpar{FIX}}
\newcommand{\new}{\marginpar{NEW}}
\newcommand{\indep}{\perp \!\!\! \perp}

\begin{abstract}
    Disentangled and invariant representations are two critical goals of representation learning and many approaches have been proposed to achieve either one of them. However, those two goals are actually complementary to each other so that we propose a framework to accomplish both of them simultaneously. We introduce a weakly supervised signal to learn disentangled representation which consists of three splits containing predictive, known nuisance and unknown nuisance information respectively. Furthermore, we incorporate contrastive method to enforce representation invariance. Experiments shows that the proposed method outperforms state-of-the-art (SOTA) methods on four standard benchmarks and shows that the proposed method can have better adversarial defense ability comparing to other methods without adversarial training.
\end{abstract}

\section{Introduction}
\label{sec:introduction}
Robust representation learning which aims at preventing overfitting and increasing generality can benefit various down-stream tasks \cite{he2015deep,bib:vae,DBLP:journals/corr/abs-1905-12506}. Typically, a DNN learns to encode a representation which contains all factors of variations of data, such as pose, expression, illumination, and age for face recognition, as well as other nuisance factors which are unknown or unlabelled. Disentangled representation learning and invariant representation learning are often used to address these challenges.

\begin{figure}[!htp]
    \centering  
    \includegraphics[width=0.8\linewidth]{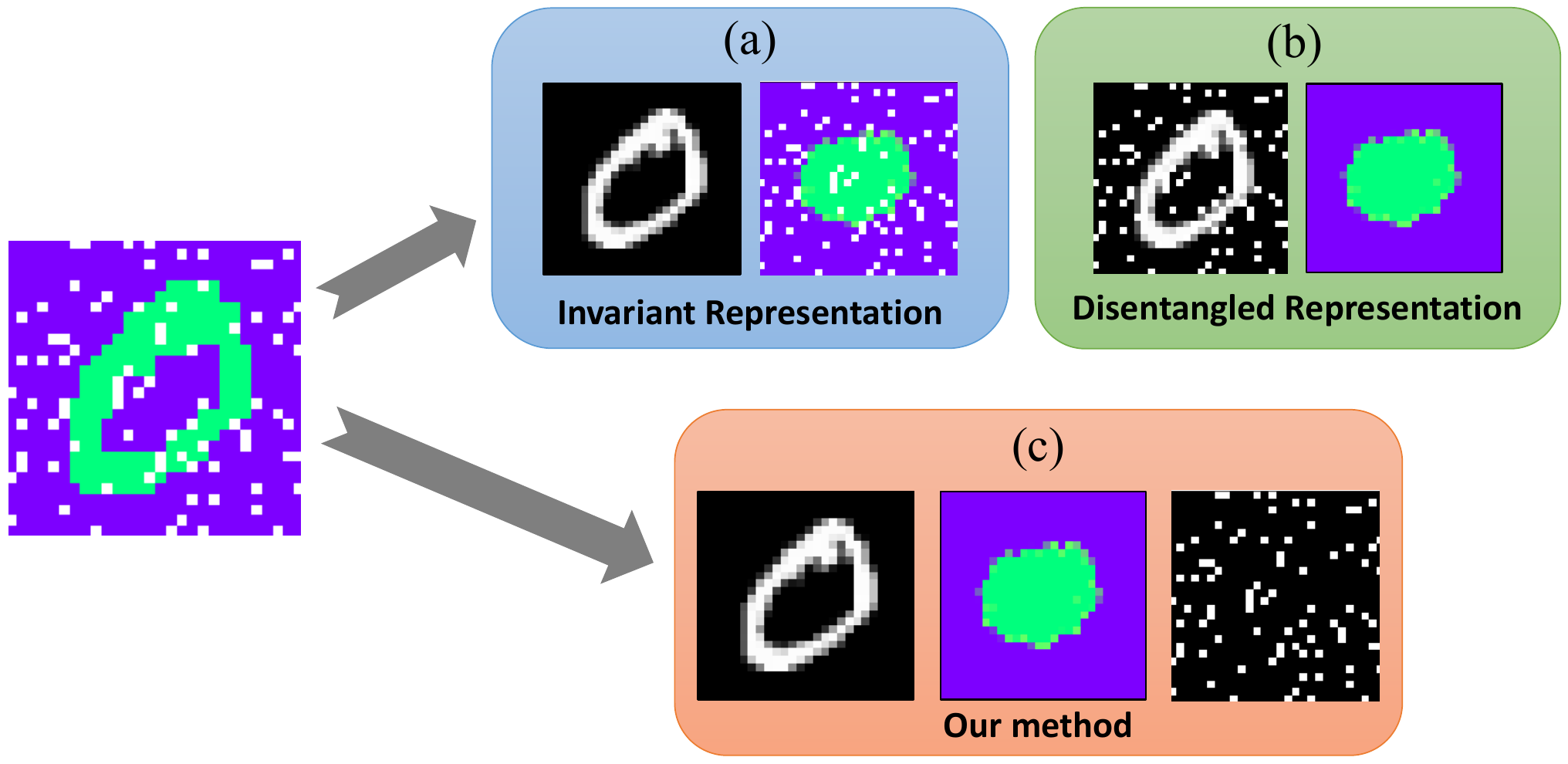}
    \caption{Given an image with nuisance factors, (a): invariant representation learning splits predictive factors from all nuisance factors; (b): disentangled representation learning splits the known nuisance factors but leaving predictive and unknown nuisance factors; (c): Our method splits predictive, known nuisance and unknown nuisance factors simultaneously. }
\label{fig:splits}
\end{figure}

For disentangled representation learning, Bengio \etal~\cite{bengio2014representation} define disentangled representation which change in a given dimension corresponding to variantion of one and only one generative factors of the input data.
Although many unsupervised learning methods have been proposed \cite{bib:betaVAE,burgess2018understanding,bib:factorvae}, Locatello \etal~\cite{locatello2019challenging} have  shown both theoretically and empirically that the factor variants disentanglement is impossible without supervision or inductive bias. To this end, recent works have adopted the concept of semi-supervised learning \cite{bib:semi-disen} and weakly supervised learning \cite{bib:adavae}. On the other hand,  Jaiswal \etal~\cite{bib:uai} take an invariant representation learning perspective in which they split representation $z$ into two parts $z = [z_p,z_n]$, where $z_p$ only contains predictive related information, and $z_n$ merely contains nuisance factors.

 Invariant representation learning aims to learn to encode predictive latent factors which are invariant to nuisance factors in inputs \cite{bib:cai,bib:uai,bib:vfae,bib:cvib,bib:irmi}. By removing information of nuisance  factors, invariant representation learning achieves good performance in challenges like adversarial attack \cite{DBLP:conf/cvpr/ChenKI20} and out-of-distribution generalization \cite{bib:uai}.  Furthermore, invariant representation learning has also been studied in the reinforcement learning settings \cite{DBLP:conf/aaai/Castro20}.

Despite the  success of either disentangled or invariant representation learning methods, the relation between these two has not been thoroughly investigated. As shown in \Cref{fig:splits}(a), invariant representation learning methods learn representations that maximize prediction accuracy by separating predictive factors from all other nuisance factors, while leaving the representations of both known and unknown nuisance factors entangled. Meanwhile, as illustrated in \Cref{fig:splits}(b), although supervised disentangled representation methods can identify known nuisance factors, it fails to handle unknown nuisance factors, which may hurt downstream prediction tasks.  Inspired by this observation, we propose a new training framework for seeking to  achieve disentanglement and invariance of representation simultaneously. To split the known nuisance factors $z_{nk}$ from predictive $z_p$ and unknown nuisance factors $z_{nu}$, we introduce the weak supervision signals to achieve disentangled representation learning. To make predictive factors $z_p$ independent of all nuisance factors $z_n$, we introduce a new invariant regularizer via reconstruction. The predictive factors from the same class are further aligned through contrastive loss to enforce  invariance. Moreover, since our model achieve more robust representation comparing to other invariant models, our model is demonstrated to obtain better adversarial defense ability.
In summary our main contributions are:

\begin{itemize}
    \item Extending and combining both disentangled and invariant representation learning and proposing a novel approach to robust representation learning. 
    \item Proposing a novel strategy for splitting the predictive, known nuisance factors and unknown nuisance factors, where mutual  independence of those factors is achieved by the reconstruction step used during training.
    \item Outperforming state-of-the-art (SOTA) models on invariance tasks on standard benchmarks. 
    \item Invariant latent representation trained using our method is also disentangled.
    \item Without using adversarial training, our model have better adversarial defense ability than other invariant models, which reflects that the generality of the model increases through our methods.
\end{itemize}

\section{Related Work}
\textbf{Disentangled representation learning}: Early works on disentangled representation learning aim at learning disentangled latent factors $z$ by implementing an autoencoder framework \cite{bib:betaVAE,bib:factorvae,burgess2018understanding}. Variational autoencoder (VAE) \cite{bib:vae} is commonly used in disentanglement learning methods as basic framework. VAE uses DNN to map the high dimension input $x$ to low dimension representation $z$. The latent representation $z$ is then mapped to high dimension reconstruction $\hat{x}$. As shown in \Cref{eq:elbo}, the overall objective function to train VAE is the evidence lower bounds (ELBO) of likelihood $\log p_{\theta}(x_1,x_2,...x_n)$, which contains two parts: quality of reconstruction and Kullback-Leibler divergence ($D_{KL}$) between distribution $q_{\phi}(z|x)$ and the assumed prior $p(z)$.  Then, VAE uses the negative of ELBO, $L_{VAE}=-ELBO$, as loss function to update the parameters in the model.
\begin{equation}
\label{eq:elbo}
\resizebox{0.9\linewidth}{!}{%
$L_{VAE}=-ELBO = -\sum_{i=1}^N \Big[ \mathbb{E}_{q_{\phi}(z|x^{(i)})}[\log p_{\theta}(x^{(i)}|z)] -D_{KL}(q_{\phi}(z|x^{(i)}||p(z)) \Big] $%
}
\end{equation}

Advanced methods based on VAE improve the disentanglement performance by implementing new disentanglement regularization. \textbf{$\beta$-VAE} \cite{bib:betaVAE} modifies the original VAE by adding a hyper-parameter $\beta$ to balance the weights of reconstruction loss and $D_{KL}$. When $\beta>1$, the model gains stronger disentanglement regularization. \textbf{AnnealedVAE} implements a dynamic algorithm to change the $\beta$ from large to small value during training. \textbf{FactorVAE} \cite{bib:factorvae} proposes to use a discriminator in order to distinguish between the joint distribution of latent factors $q(z)$ and multiplication of marginal distribution of every latent factor $\prod q(z_i)$. By using the discriminator,  \textbf{FactorVAE} can automatically finds a better balance between reconstruction quality and disentangled representation. Compared to \textbf{$\beta$-VAE},  \textbf{DIP-VAE} \cite{bib:dipvae} adds another regularization $D(q_{\phi}(z) || p(z))$ between the marginal distribution of latent factors $q_{\phi}(z) = \int q_{\phi}(z|x)p(x) dx$ and the prior $p(z)$ to further aid disentangled representation learning, where $D$ can be any proper distance function such as mean square error. \textbf{$\beta$-TCVAE} proposed by \cite{chen2019isolating} modifies the $D_{KL}$ used in \textbf{VAE} into three part: \emph{total correlation, index-coded mutual information} and \emph{ dimension-wise KL divergence}. To overcome the challenge proposed by \cite{locatello2019challenging},  \textbf{AdaVAE} \cite{bib:adavae} purposely chooses pairs of inputs as supervision signal to learn representation disentanglement.

\begin{figure}[]
    \centering
    \includegraphics[width=\textwidth]{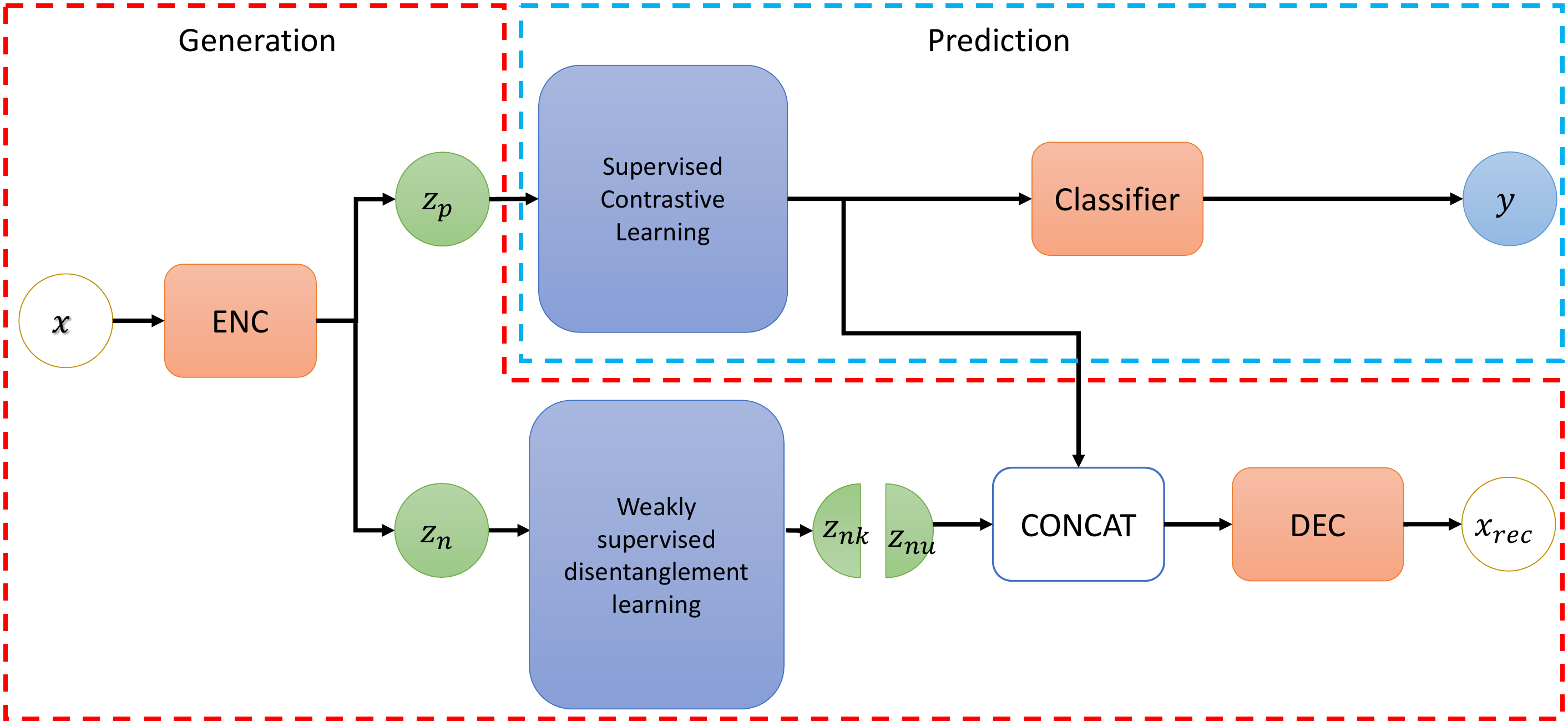}
    \caption{Architecture of the model. \textcolor{red}{Red box} is the generation part and the \textcolor{blue}{blue box} is the prediction part. Generation module aims at learning and splitting known nuisance factors $z_{nk}$ and $z_{nu}$, and the prediction module aims at learning good predictive factors $z_p$. }
    \label{fig:architecture}
\end{figure}
\textbf{Invariant Representation Learning}: The methods that aim at learning invariant representation can be classified into two groups: those methods that require annotations of nuisance factors \cite{bib:nnmmd,bib:vfae} and those that do not.  A considerable number of approaches using nuisance factors annotations have been recently proposed. By implementing a regularizer which minimizes the Maximum Mean Discrepancy (MMD) on neural network (NN), The \textbf{NN+MMD} approach \cite{bib:nnmmd} removes affects of nuisance from predictive factors. On the basis of \textbf{NN+MMD}, The Variational Fair Autoencoder (\textbf{VFAE}) \cite{bib:vfae} uses special priors which encourage independence between nuisance factors and ideal invariant factors. The Controllable Adversarial Invariance (\textbf{CAI}) \cite{bib:cai} approach applies the gradient reversal trick \cite{bib:dann} which penalizes the model if latent representation has information of nuisance factors. \textbf{CVIB} \cite{bib:cvib} proposes a conditional form of Information Bottleneck (IB) and encourages the invariant representation learning by optimizing its variational bounds.

However, due to the constrains of demanding annotations, those methods take more effort to pre-process the data and encounter challenges when the annotations are inaccurate or insufficient.  Comparing to annotation-eager approaches, annotation-free methods are easier to be implemented in practice. The Unsupervised Adversarial Invariance (\textbf{UAI}) \cite{bib:uai} splits the latent factors into factors useful for prediction and nuisance factors. \textbf{UAI} encourages the independence of those two latent factors by incorporating competition between the prediction and the reconstruction objectives. \textbf{NN+DIM} \cite{bib:irmi} achieves invariant representation by using pairs of inputs and applying a neural network based mutual information estimator to minimize the mutual information between two shared representations. Furthermore, Sanchez \etal~\cite{bib:irmi} employ a discriminator to distinguish the difference between shared representation and nuisance representation.

\section{Learning disentangled and invariant representation}

\subsection{Model Architecture}
\begin{figure}[]
    \centering
    \includegraphics[width=\textwidth]{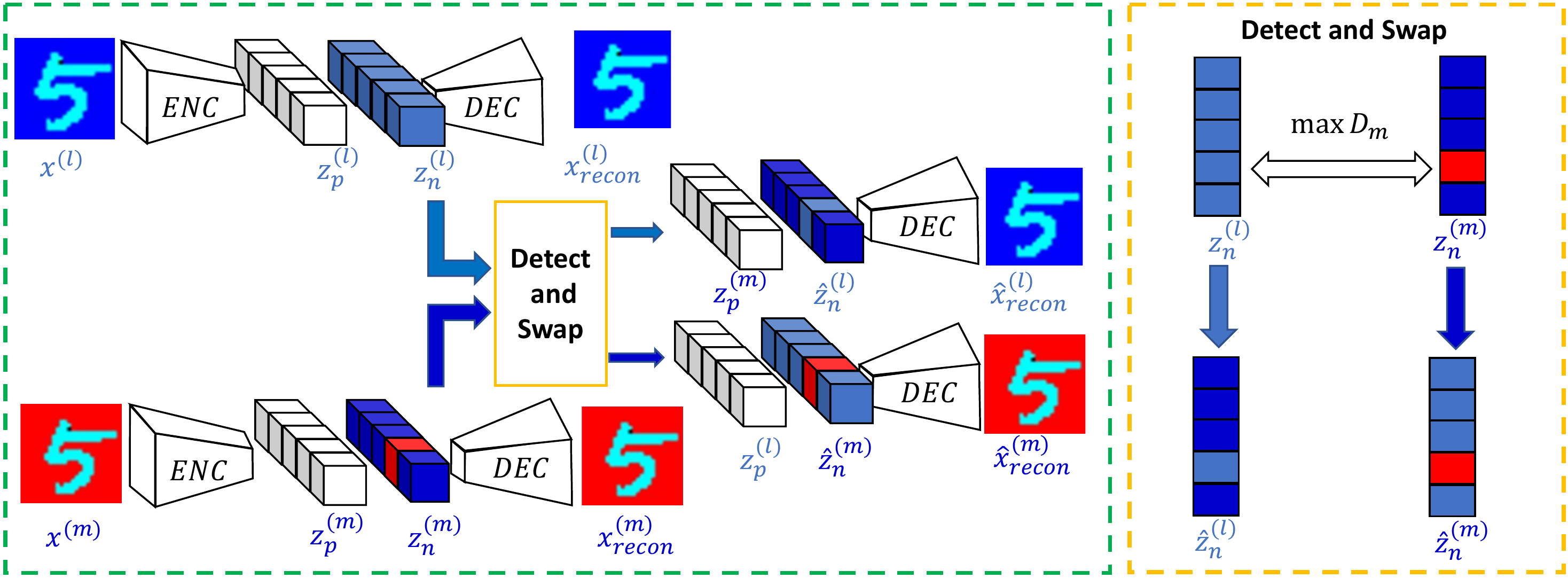}
    \caption{Weakly supervised disentanglement representation learning for known nuisance factors $z_{nk}$}
    \label{fig:disentanglement_weak}
\end{figure}

As illustrated in \Cref{fig:architecture}, the architecture of the proposed model contains two components: a generation module and a prediction module. Similar to VAE, the generation module performs the encoding-decoding task. However, it encodes the input $x$ into latent factors $z$, $z =[z_p,z_n]$, where $z_p$ represents the latent predictive factors that contains useful information for the prediction task, whereas $z_n$ represents the latent nuisance factors and can be further divided into known latent factors $z_{nk}$ and unknown nuisance factors $z_{nu}$.

 $z_{nk}$ are discovered and separated from $z_n$ via weakly supervised disentangled representation learning, where the joint distribution $p(z_{nk}) = \prod_{i} p(z_{{nk}_i})$. Since $z_n$ is the split containing nuisance factor, after $z_{nk}$ is identified, the remaining factors of $z_n$ naturally result in unknown nuisance factors $z_{nu}$. Then, $z_p$ and $z_n$ are concatenated for generating reconstructions $x_{rec}$ which are used to measure the quality of reconstruction. To enforce the independence between $z_p$ and $z_n$, we add a regularizer using another reconstruction task, where the average mean and variance of predictive factors $z_p$ are used to form new latent factors $\Bar z_p$ and it will be discussed in \Cref{sec:invariant}. In the prediction module, we further incorporate contrastive loss to cluster the predictive latent factors belonging to the same class.

\subsection{Learning independent known nuisance factors $z_{nk}$}
\label{sec:weakly_disentanglement}

As illustrated in \Cref{fig:architecture}, the known nuisance factors $z_{nk}$ are discovered and separated  from $z_n$, where $p(z_{nk}) = \prod_i p(z_{nk_i})$, since nuisance information is expected to be present only within $z_n$.

To fulfill the theoretical requirement of including supervision signal for disentangled representation learning as proven in \cite{locatello2019challenging}, we use selected pairs of inputs $x^{(l)}$ and $x^{(m)}$ as supervision signals, where only a few common generative factors are shared.
As illustrated in \Cref{fig:disentanglement_weak}, during training, the network encodes a pair of inputs $x^{(l)}$ and $x^{(m)}$ into two latent factors $z^{(l)} = [z^{(l)}_p, z^{(l)}_n]$ and $z^{(m)} = [z^{(m)}_p, z^{(m)}_n]$ respectively, which are then decoded to reconstruct $x^{(l)}_{rec}$ and $x^{(m)}_{rec}$. To encourage representation disentanglement, certain elements of $z_n^{(l)}$ and $z_n^{(m)}$ are \emph{detected and swapped} to generate two new corresponding latent factors $\hat z^{(l)}$ and $\hat z^{(m)}$. 
\begin{figure}[]
    \centering
    \includegraphics[width=0.8\textwidth]{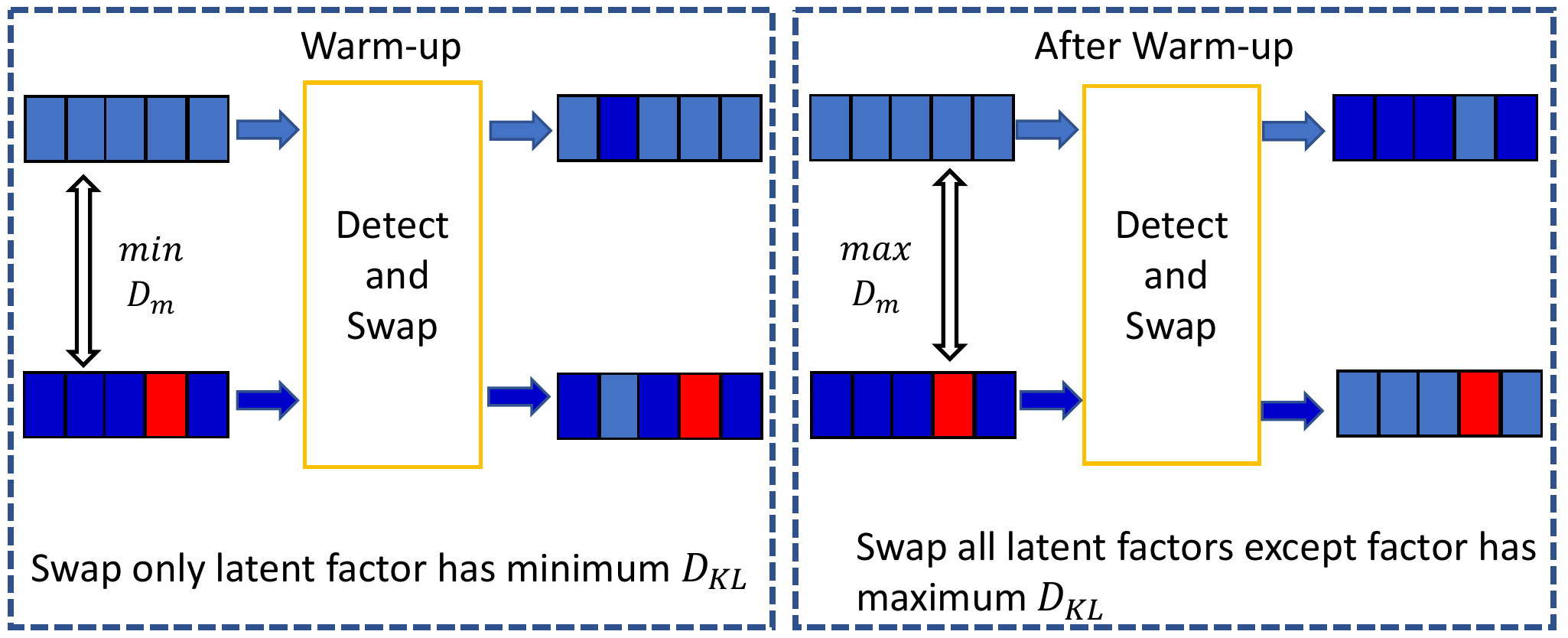}
    \caption{In early training stages, small number of latent factors are swapped. the number of latent factors to be swapped increases gradually.}
    \label{fig:first_strategy}
\end{figure}
The two new latent factors are then decoded to new reconstructions $\hat x^{(l)}_{rec}$ and $\hat x^{(m)}_{rec}$. By comparing $\hat x_{rec}$ with $x_{rec}$, the known nuisance factors $z_{nk}$ are discovered and the elements of $z_{nk}$ are enforced to be mutually independent with each other.

\paragraph{Selecting image pairs for training and latent factors assumptions:}
\label{pargraph:assumptions}
 As mentioned by \cite{bib:betaVAE}, the true world simulator using generative factors to generate $x$ can be modeled as: $p(x|v,w) = Sim(v,w)$, where $v$ is the generative factors and $w$ is other nuisance factors. Inspired by this, we choose pairs of images by  randomly selecting several generative factors to be the same and keeping the value of other generative factors to be random. Each image $x$ has corresponding generative factors $v$, and the training pair is generated as follows: we first randomly select a sample $x^{(l)}$ whose generative factors are $v^{(l)}=[v_1,v_2,...v_n]$. We then randomly change the value of $k$ elements in $v^{(l)}$ to form a new generative factors $v^{(m)}$ and choose another sample $x^{(m)}$ according to $v^{(m)}$. During training, indices of different generative factors between $v^{(l)}$ and $v^{(m)}$, and the groundtruth value of all generative factors are not available to the model. The model is weakly supervised since it is trained with only the knowledge of the number of factors $k$ that have changed.  Ideally, if the model can learn a disentangled representation, the model will encode the image pair $x^{(l)}$ and $x^{(m)}$ to the corresponding representations $z^{(l)}$ and $z^{(m)}$ which have the characteristic shown in \Cref{eq:pairs}. We annotate the set of all different elements between $z^{(l)}$ and $z^{(m)}$ to be $df_z$ and set of all latent factors to be $d_z$ such that $df_z \subseteq d_z$.
\begin{equation}
\begin{split}
\label{eq:pairs}
    p(z_{n_j}^{(l)}|\mathbf{x^{(l)}}) &= p(z_{n_j}^{(m)}|\mathbf{x^{(m)}}); ~j \notin df_z \\
    p(z_{n_i}^{(l)}|\mathbf{x^{(l)}}) &\neq p(z_{n_i}^{(m)}|\mathbf{x^{(m)}}); ~i \in df_z 
\end{split}
\end{equation}

\paragraph{Detecting and Swapping the distinct latent factors:}
\label{pargraph:ds}
VAE adopts the reparameterization to make posterior distribution $q_{\theta}(z|x)$ differentiable, where the posterior distribution of latent factors is commonly assumed to be a factorized multivariate Gaussian: $p(z|x) = q_{\theta}(z|x)$ \cite{bib:vae}. By this assumption, we can directly measure the mutual information between the corresponding dimensions of the two latent representations $z^{(l)}$ and $z^{(m)}$ by  measuring the divergence ($D_m$), which can be KL divergence ($D_{KL}$). We show the process of detecting distinct latent factors in \Cref{eq:latentkl}, where a larger value of $D_{KL}$ implies higher difference between the  two corresponding latent factor distributions. 
\begin{small}
\begin{align}
    \label{eq:latentkl}
    D_{KL}(q_{\phi}(z_i^{(l)}|x^{(l)})||q_{\phi}(z_i^{(m)}|x^{(m)})) 
      =\frac{({\sigma^{(l)}_i})^2 + (\mu^{(l)}_i - \mu^{(m)}_i)^2}{2{(\sigma^{(l)}_i})^2} + log(\frac{\sigma^{(l)}_i}{\sigma^{(m)}_i}) -\frac{1}{2} 
   \end{align}
\end{small}

Since the model only has the knowledge of the number of different generative factors $k$, we swap all corresponding dimension elements of $z_n^{(l)}$ and $z_n^{(m)}$ except the top $k$ highest $D_{m}$ value elements. We incorporate this swapping step to create two new latent representations $\hat z_n^{(l)}$ and $\hat z_n^{(m)}$ shown in \Cref{eq:exchange}.
\begin{equation}
\begin{split}
    \label{eq:exchange}
    \hat z_{n_i}^{(l)} & = z_{n_i}^{(m)};  
    \hat z_{n_i}^{(m)}  = z_{n_i}^{(l)};  ~ i \notin df_z \\
    \hat z_{n_j}^{(l)} & = z_{n_j}^{(l)};   
    \hat z_{n_j}^{(m)}  = z_{n_j}^{(m)};   ~ j \in df_z \\ 
\end{split}
\end{equation}
\paragraph{Disentangled representation loss function:}
After  $\hat z_n^{(l)}$ and $\hat z_n^{(m)}$ are obtained, they are concatenated with $z_p^{(m)}$ and $z_p^{(l)}$ respectively, to generate two new latent representations $\hat z^{(l)}=[z_p^{(m)},\hat z_n^{(l)}]$ and $\hat z^{(m)}=[z_p^{(l)},\hat z_n^{(m)}]$. $\hat z^{(l)}$ and $\hat z^{(m)}$ are decoded into new reconstructions $\hat x^{(l)}$ and $\hat x^{(m)}$. 
Since there are only $k$ different generative factors between pair of images, ideally, after encoding the images, there should also be merely $k$ pairs of different distributions on the latent representation space.
By swapping other latent factors except them, the new representations $\hat z^{(l)}$ and $\hat z^{(m)}$ are the same with the original representations $z^{(l)}$ and $z^{(m)}$. 
Accordingly, the new reconstructions  $\hat x_{rec}^{(l)}$ and $\hat x_{rec}^{(m)}$ should be identical to the original reconstruction  $x_{rec}^{(l)}$ and $x_{rec}^{(m)}$. Therefore, we design the disentangled representation loss in \Cref{eq:overall_loss}, where $D$ can be any suitable distance function \eg~mean square error (MSE) or binary cross-entropy (BCE). 
\begin{equation}
    \label{eq:overall_loss}
    L =L_{VAE}(x^{(l)}_{rec},z^{(l)}) + L_{VAE}(x^{(m)}_{rec},z^{(m)} )+ D(\hat{x}^{(l)}_{rec}, x^{(l)}_{rec}) + D(\hat{x}^{(m)}_{rec}, x^{(m)}_{rec})
\end{equation}
\paragraph{Training Strategies for disentangled representation learning:} 
To further improve the performance of disentangled representation learning, we design two strategies: \emph{warmup by amount} and \emph{warmup by difficulty}. Recalling that in the swapping step, the model needs to swap $|d_z| - k$ elements of latent representations. At beginning,  exchanging too many latent factors will easily lead to mistakes. Therefore, in the first strategy, we gradually increase the number of latent factors being swapped from $1$ to $|d_z|-k$. Further, to smoothly increase the training difficulty, we set the number of different generative factors to be $1$ at the beginning and increase the number of different generative factors as training continues.

\subsection{Learning invariant predictive factors $z_p$}
\label{sec:invariant}
After we obtain the disentangled representation $z_{nk}$, the predictive factors $z_p$ may still be entangled with $z_{nu}$. Therefore, we need to add other constraints to achieve fully invariant representation of $z_p$.

\paragraph{Making $z_p$ independent of $z_n$:} 
As shown in \cite{locatello2019challenging}, supervision signals need to be introduced for disentangled representation. Similarly, the independency of $z_p$ and $z_n$ also needs the help from a supervision signals as we discuss in Appendix. Luckily, for supervised training, a batch of samples naturally contains supervision signal. Similar to \Cref{eq:pairs}, the distribution of the representations $z_p$ should be the same for the same class and can be shown in \Cref{eq:zp_disentangle} where $C(x^{(l)})$ means the class of sample $x^{(l)}$.
\begin{equation}
\begin{split}
\label{eq:zp_disentangle}
    p(z_p^{(l)}|\mathbf{x^{(l)}}) &= p(z_p^{(m)}|\mathbf{x^{(m)}}); ~ C(x^{(l)})= C(x^{(m)}) \\
    p(z_p^{(l)}|\mathbf{x^{(l)}}) &\neq p(z_p^{(m)}|\mathbf{x^{(m)}}); ~C(x^{(l)})\neq C(x^{(m)}) 
\end{split}
\end{equation}
Similar to the method we use for disentangled representation learning, we generate a new latent representation $\Bar z_p$ and its corresponding reconstruction $\Bar x_{rec-p}$. Then, we enforce the disentanglement between $z_p$ and $z_n$ by comparing the new reconstruction $\Bar x_{rec-p}$ and $x_{rec}$. In contrast to the swapping method mentioned in \Cref{sec:weakly_disentanglement}, since the batch of samples used for training often contains more than two samples from the same class, the swapping method is hard to be implemented in this situation. Therefore, we generate the new latent representations $\Bar z_p$ by calculating the average mean $\Bar \mu_p$ and average variance $\Bar V_p$ of the latent representations from the same class as shown in \Cref{eq:mean_zp}.
\begin{equation}
\begin{split}
\label{eq:mean_zp}
   &\Bar{z_p} = \mathcal{N}(\Bar{\mu_p}, \Bar{V_p}); ~\Bar x_{rec-p} = Decoder([\Bar z_p,z_n]) \\ 
    ~ \Bar{\mu_p} = &\frac{1}{|C|} \sum \mu^{(i)}_p ~;~ 
    \Bar{V_p} = \frac{1}{|C|} \sum V^{i}_p;~ where~ \forall i \in C
\end{split}
\end{equation}
We then generate the new reconstruction $\Bar x_{rec-p}$ using the same decoder as in other reconstruction tasks and enforce the disentanglement of $z_p$ and $z_n$ by calculating the $D({x_{rec}, \Bar x_{rec_p}})$ and update the parameters of the model according to its gradient.

\paragraph{Constrastive feature alignment:}
To achieve invariant representation, we need to make sure the latent representation that is useful for prediction can also be clustered according to their corresponding classes. Even though the often  used cross-entropy (CE) loss can accomplish similar goals, the direct goal of CE loss is to achieve logit-level alignment and change the representations distribution according to the logits, which does not guarantee the uniform distribution of features. Alternatively, we incorporate contrastive methods to ensure that representation/feature alignment can be accomplished effectively \cite{Wang_2021_CVPR}.  

Similar to  \cite{bib:super_contras}, we use supervised contrastive loss to achieve feature alignment and cluster the representations $z_p$ according to their classes as shown in  \Cref{eq:sup_contrast} where $C$ is the set that contains samples from the same class and $y_p = y_i$.
\begin{equation}
\label{eq:sup_contrast}
  \mathcal{L}_{sup}
  =\sum_{i\in I}\frac{-1}{|C|}\sum_{p\in C}\log{\frac{\text{exp}\left(z_i \cdot z_p/\tau\right)}{\sum\limits_{a\in A(i)}\text{exp}\left(z_i\cdot z_a/\tau\right)}}
\end{equation}

 The final loss function used to train the model, after adding the standard cross-entropy(CE) loss to train the classifier, is given by \Cref{eq:all_loss}.
\begin{equation}
\begin{aligned}
\label{eq:all_loss}
    &L = L_{CE}(x,y) + L_{VAE} + \alpha L_{disentangle} + \beta L_{Sup} + \gamma L_{Z_p} \\
    &L_{disentangle} = D(\hat{x}^{(l)}_{rec}, x^{(l)}_{rec}) + D(\hat{x}^{(m)}_{rec}, x^{(m)}_{rec}) \\
    &L_{Z_p} = D(\Bar x_{rec-p}, x_{rec} )
\end{aligned}
\end{equation}

\begin{table*}[]
\caption{Test average and worst accuracy results on Colored-MNIST, 3dShapes and MPI3D. \textbf{Bold, Black}: best result}
\label{table:invariant1}
\centering
\renewcommand{\arraystretch}{1.2}
{
\begin{adjustbox}{width=0.93\textwidth}
\begin{tabular}{c |c c |c c |c c }
\hlineB{2}
                & \multicolumn{2}{c}{\textbf{Colored-MNIST}} & \multicolumn{2}{c}{\textbf{3dShapes}}                                                                 & \multicolumn{2}{c}{\textbf{MPI3D}}                                                                   \\
\multirow{-2}{*}{\textbf{Models}} & \textbf{Avg Acc}               &  \textbf{Worst Acc}               & \multicolumn{1}{c}{\textbf{Avg Acc}} & \multicolumn{1}{c|}{\textbf{Worst Acc}} & \multicolumn{1}{c}{\textbf{Avg Acc}} & \multicolumn{1}{c}{\textbf{Worst Acc}} \\ \cmidrule(r){1-1} \cmidrule(r){2-3} \cmidrule(r){4-5} \cmidrule(r){6-7}
Baseline                                                  & 95.12 $\pm$ 2.42                        & 66.17 $\pm$ 3.31                          & \textbf{98.87 $\pm$ 0.52}                                                    & 96.89 $\pm$ 1.25                                                     & 90.12  $\pm$ 3.13                                                    & 87.89  $\pm$ 4.31                                                      \\

VFAE~\cite{bib:vfae}                & 93.12 $\pm$ 3.07                         & 65.54 $\pm$ 6.21                          & 97.72 $\pm$ 0.81           & 93.34 $\pm$ 1.05                                                          & 86.69 $\pm$ 3.12                                                       & 82.43 $\pm$ 3.25                                                         \\
CAI~\cite{bib:cai}     & 93.56 $\pm$ 2.76          & 63.17 $\pm$ 5.61                           & 97.62 $\pm$ 0.53                      & 94.32 $\pm$ 0.89                                             & 86.63 $\pm$ 2.14                         & 82.16 $\pm$ 5.83                                                          \\
CVIB~\cite{bib:cvib}    & 93.31 $\pm$ 3.09          & 70.12 $\pm$ 4.77              & 97.11 $\pm$ 0.59                                  & 94.46 $\pm$ 0.90                                                          & 87.04 $\pm$ 3.02                          & 85.61 $\pm$ 2.08                                                         \\
UAI~\cite{bib:uai}     & 94.74 $\pm$ 2.19                         &  74.25 $\pm$ 2.69                           & 97.13 $\pm$ 1.02                     & 95.21 $\pm$ 1.03                                              & 87.89 $\pm$ 4.23                      & 83.01 $\pm$ 2.21                                                         \\
NN+DIM~\cite{bib:irmi}   & 94.48 $\pm$ 2.35         & 80.25$\pm$ 3.44           & 97.03 $\pm$ 1.07                      & 96.02 $\pm$ 0.46   &  88.81 $\pm$ 1.37                                                       & 82.01 $\pm$ 3.34                                                         \\ \hdashline
\textbf{Our model}   & \textbf{97.96 $\pm$ 1.21}                         & \textbf{90.43$\pm$ 2.79}              & 98.52 $\pm$ 0.51 & \textbf{97.63 $\pm$ 0.72}                                                          & \textbf{91.32 $\pm$ 2.38}                                                       & \textbf{89.17$ \pm$ 2.69}                                                         \\ \hlineB{2}
\end{tabular}
\end{adjustbox}
}
\end{table*}
\section{Experiments Evaluation}
\subsection{Benchmarks, Baselines and Metrics}
\label{sec:bench}
The main objective of this work is to learn invariant representations and reduce overfitting to nuisance factors. Meanwhile, as a secondary objective, we also want to ensure that the learned representations are at least not less robust to adversarial attacks. Therefore, all models are evaluated on both invariant representation learning task and adversarial robustness task. We use four (4) dataset with different underlying factors of variations to evaluate the model:

\begin{itemize}
    \item \textbf{Colored-MNIST} Colored-MNIST dataset is augmented version of MNIST \cite{lecun-mnisthandwrittendigit-2010} with two known nuisance factors: digit color and background color \cite{bib:irmi}. During training, the background color is chosen from three (3) colors and digit color is chosen from other six (6) colors. In test, we set the background color into three (3) new colors which is different from training set. 

   \item \textbf{Rotation-Colored-MNIST} This dataset is further augmented version of Colored-MNIST. The background color and digit color setting is the same with the Colored-MNIST. This dataset further contains digits rotated to four (4) different angles $ \Theta_{train} = \{0, \pm 22.5, \pm 45\}$. For test data, the rotation angles for digit is set to $\Theta_{test} = \{0, \pm 65, \pm 75\}$. The rotation angles are used as unknown nuisance factors.

   \item \textbf{3dShapes} \cite{3dshapes18} contains 480,000 RGB $64\times64\times3$ images and the whole dataset has six (6) different generative factors. We choose object shape (four (4) classes) as the prediction task and only half number of object colors are used during training, and the remaining half of object color samples are used to evaluate performance of invariant representation. 

    \item \textbf{MPI3D} \cite{gondal2019transfer} is a real-world dataset contains 1,036,800 RGB images and the whole dataset has seven (7) generative factors. Like 3dShapes, we choose object shape (six (6) classes) as the prediction target and half of object colors are used for training. 
\end{itemize}

\begin{table*}[]
\caption{Test average accuracy and worst accuracy results on Rotation-Colored-MNIST with different rotation angles.  \textbf{Bold, Black}: best result}
\label{table:invariant2}
\centering
\renewcommand{\arraystretch}{1.2}
{
\begin{adjustbox}{width=\textwidth}
\begin{tabular}{c|cc|cc|cc|cc}
\hlineB{2}
\multirow{3}{*}{\textbf{Models}} & \multicolumn{8}{c}{\textbf{Rotation-Colored-MNIST}}                                                                                                                                                                                                                      \\
            & \textbf{Avg Acc} & \textbf{Worst Acc} & \multicolumn{1}{c}{\textbf{Avg Acc}} & \multicolumn{1}{c|}{\textbf{Worst Acc}} & \multicolumn{1}{c}{\textbf{Avg Acc}} & \multicolumn{1}{c|}{\textbf{Worst Acc}} & \multicolumn{1}{c}{\textbf{Avg Acc}} & \multicolumn{1}{c}{\textbf{Worst Acc}}          \\
                & \multicolumn{2}{c|}{\textbf{-75}}      & \multicolumn{2}{c|}{\textbf{-65}}                                              & \multicolumn{2}{c|}{\textbf{+65}}   & \multicolumn{2}{c}{\textbf{+75}}                                   \\  \cmidrule(r){1-1} \cmidrule(r){2-3} \cmidrule(r){4-5} \cmidrule(r){6-7} \cmidrule{8-9}
Baseline& 77.0 $\pm$1.3            & 62.3 $\pm$1.9              & 89.7 $\pm$1.2         & 77.5 $\pm$2.2                                  & 85.8 $\pm$1.2        &  65.8 $\pm$3.0                                 & 68.3 $\pm$2.2                                & 49.9 $\pm$4.6                             \\
VFAE~\cite{bib:vfae}    & 72.2 $\pm$2.4     & 58.9 $\pm$2.3     & 85.8 $\pm$1.7 & 74.4 $\pm$2.5 & 84.1 $\pm$2.1                                & 64.6 $\pm$3.7          & 71.7 $\pm$1.3                                & 48.0 $\pm$3.8                       \\
CAI~\cite{bib:cai}    & 74.9 $\pm$0.9           & 59.3 $\pm$3.9              & 86.5 $\pm$1.9      & 77.3 $\pm$2.0 & 84.2 $\pm$1.7             & 67.8 $\pm$1.9         & 64.7 $\pm$4.2            & 42.9 $\pm$3.7                           \\
CVIB~\cite{bib:cvib}    & 76.1 $\pm$0.8            & 59.2 $\pm$3.0              & 88.6 $\pm$0.9     & 79.1 $\pm$1.2                             & 85.6 $\pm$0.7             & 68.8 $\pm$2.9         & 72.2 $\pm$1.2         & 53.4 $\pm$ 2.6                               \\
UAI~\cite{bib:uai}  & 76.0 $\pm$1.7            & 61.1 $\pm$5.6              & 88.8 $\pm$0.7    & 80.0 $\pm$0.9                                  & 85.4 $\pm$1.6    & 68.2 $\pm$2.3         & 70.2 $\pm$ 0.9                               & 51.1 $\pm$2.3                             \\
NN+DIM~\cite{bib:irmi}  & 77.6 $\pm$2.6            & 69.2 $\pm$2.7             & 85.2 $\pm$3.4                                & 76.3 $\pm$4.3    & 84.6 $\pm$3.1    & 66.7 $\pm$3.7         & 68.4 $\pm$3.1         & 53.2 $\pm$5.6                        \\ \hdashline
\textbf{Our model}       & \textbf{81.0 $\pm$2.1}            & \textbf{75.3 $\pm$2.5}              & \textbf{90.8 $\pm$1.6}                & \textbf{85.7 $\pm$2.4}      & \textbf{87.3 $\pm$2.5}    & \textbf{82.3 $\pm$2.1}                                  & \textbf{73.2 $\pm$2.3}              & \textbf{63.3 $\pm$2.9}                                  \\ \hlineB{2}
\end{tabular}
\end{adjustbox}
}

\end{table*}

Prediction accuracy is used to evaluate the performance of invariant representation learning.  Furthermore, we record both average test accuracy and worst-case test accuracy which was suggested by \cite{Sagawa*2020Distributionally}. We find that using \Cref{eq:all_loss} directly does not guarantee good performance. This may be caused by inconsistent behavior of CE loss and supervised contrastive loss. Thus, we separately train the classifier using CE loss and use remaining part of total loss to train the rest of the model.

Meanwhile, the performance of representation disentanglement is also important for  representation invariance since it can evaluate the invariance of latent factors representing known nuisance factors.

We adopt the following metrics to evaluate the performance of disentangled representation. All metrics range from $0$ to $1$, where $1$ indicates that the latent factors are fully disentangled --- (1) \textbf{Mutual Information Gap (MIG)} \cite{chen2019isolating} evaluates the gap of top two highest mutual information between a latent factors and generative factors. 
(2) \textbf{Separated Attribute Predictability (SAP)} \cite{bib:dipvae} measures the mean of the difference of perdition error between the top two most predictive latent factors.
(3) \textbf{Interventional Robustness Score (IRS)} \cite{suter2019robustly} evaluates reliance of a latent factor solely on generative factor regardless of other generative factors.
(4) \textbf{FactorVAE (FVAE) score} \cite{bib:factorvae} implements a majority vote classifier to predict the index of a fixed generative factor and take the prediction accuracy as the final score value. 
(5) \textbf{DCI-Disentanglement (DCI)} \cite{bib:dci} calculates the entropy of the distribution obtained by normalizing among each dimension of the learned representation for predicting the value of a generative factor.

\subsection{Comparison with Previous Work}
\label{sec:quantitative}

\begin{table*}[]
\caption{Disentanglement metrics on 3dShapes and MPI3D. \textbf{Bold, Black}: best result}
\label{table:disentangle2}
\centering

\renewcommand{\arraystretch}{1.2}
{
\begin{adjustbox}{width=0.9\textwidth}
\begin{tabular}{ccccccccccc}
\hlineB{2}
\multicolumn{1}{c|}{}                                  & \multicolumn{5}{c|}{\textbf{3dShapes}}                                                                                           & \multicolumn{5}{c}{\textbf{MPI3D}}                                        \\
\multicolumn{1}{c|}{\multirow{-2}{*}{\textbf{Models}}} & \textbf{MIG}                  & \textbf{SAP}                  & \textbf{IRS} & \textbf{FVAE} & \multicolumn{1}{c|}{\textbf{DCI}} & \textbf{MIG} & \textbf{SAP} & \textbf{IRS} & \textbf{FVAE} & \textbf{DCI} \\ \hline
\multicolumn{11}{c}{\textbf{Unsupervised Disentanglement Leanring}}                                                                                                                                                                                                   \\ \hline
\multicolumn{1}{c|}{$\beta$-VAE~\cite{bib:betaVAE} }                           & 0.194                        & 0.063                        & 0.473       & 0.847        & \multicolumn{1}{c|}{0.246}       & 0.135        & 0.071        & 0.579        & 0.369         & 0.317        \\
\multicolumn{1}{c|}{AnnealedVAE~\cite{burgess2018understanding}}                       & 0.233                        & 0.087                        & 0.545       & 0.864        & \multicolumn{1}{c|}{0.341}       & 0.098        & 0.038        & 0.490        & 0.397         & 0.228        \\
\multicolumn{1}{c|}{FactorVAE~\cite{bib:factorvae}}                         & 0.224                        & 0.0440                        & 0.630       & 0.792        & \multicolumn{1}{c|}{0.304}       & 0.092        & 0.031        & 0.529        & 0.379         & 0.164        \\
\multicolumn{1}{c|}{DIP-VAE-I~\cite{bib:dipvae}}                         & 0.143                        & 0.026                        & 0.491       & 0.761        & \multicolumn{1}{c|}{0.137}       & 0.104        & 0.073        & 0.476        & 0.491         & 0.223        \\
\multicolumn{1}{c|}{DIP-VAE-II~\cite{bib:dipvae}}                        & 0.137                        & 0.020                        & 0.424       & 0.742        & \multicolumn{1}{c|}{0.083}       & 0.131        & 0.075        & 0.509        & 0.544         & 0.244        \\
\multicolumn{1}{c|}{$\beta$-TCVAE~\cite{chen2019isolating}}                        & 0.364                        & 0.096                        & 0.594       & 0.970        & \multicolumn{1}{c|}{0.601}       & 0.189        & 0.146        & \textbf{0.636}        & 0.430         & 0.322        \\ \hline
\multicolumn{11}{c}{\textbf{Weakly-Supervised Disentanglement Learning}}                                                                                                                                                                                              \\ \hline
\multicolumn{1}{c|}{Ada-ML-VAE~\cite{bib:adavae}}                        & 0.509                        & 0.127                        & 0.620       & 0.996        & \multicolumn{1}{c|}{0.940}       & 0.240        & 0.074        & 0.576        & 0.476         & 0.285        \\
\multicolumn{1}{c|}{Ada-GVAE~\cite{bib:adavae}}                          & 0.569                        &  0.150 & 0.708       & 0.996        & \multicolumn{1}{c|}{\textbf{0.946}}       & 0.269        & 0.215        & 0.604        & 0.589         & 0.401        \\ \hdashline
\multicolumn{1}{c|}{\textbf{Our model}}                         &  \textbf{0.716} & \textbf{0.156}                        & \textbf{0.784}       & 0.996        & \multicolumn{1}{c|}{0.919}       & \textbf{0.486}        & \textbf{0.225}        & 0.615        & \textbf{0.565}         & \textbf{0.560}        \\ \hlineB{2}
\end{tabular}
\end{adjustbox}
}

\end{table*}

We show invariance learning results which are the test average accuracy and worst accuracy in \Cref{table:invariant1,table:invariant2}.  For Color-Rotation-MNIST dataset, since we rotate the test samples with $\theta \in \Theta_{test} = \{ \pm 65, \pm 75\}$ and those angles are different with training rotation angles $\theta \in \Theta_{train} = \{0, \pm 22.5, \pm 45\}$, we record each average accuracy and worst accuracy under each rotation angles. The baseline model is the regular VGG16 model with no extra components for representation invariance. Our model largely outperforms prior work. 

To compare the performance of disentanglement, we show the results of disentanglement representation learning in \Cref{table:disentangle2}. Since the Color-MNIST and Rotation-Color-MNIST have only two generative factors, models for disentanglement learning tends to achieve nearly perfect disentanglement metric scores, which makes the results seems trivial.
Therefore, we only record the disentanglement scores tested on \textbf{3dshapes} and \textbf{MPI3D} datasets. 

\begin{figure}[]
    \centering %
\begin{subfigure}{0.33\textwidth}
  \includegraphics[width=\linewidth,height=0.6\textwidth]{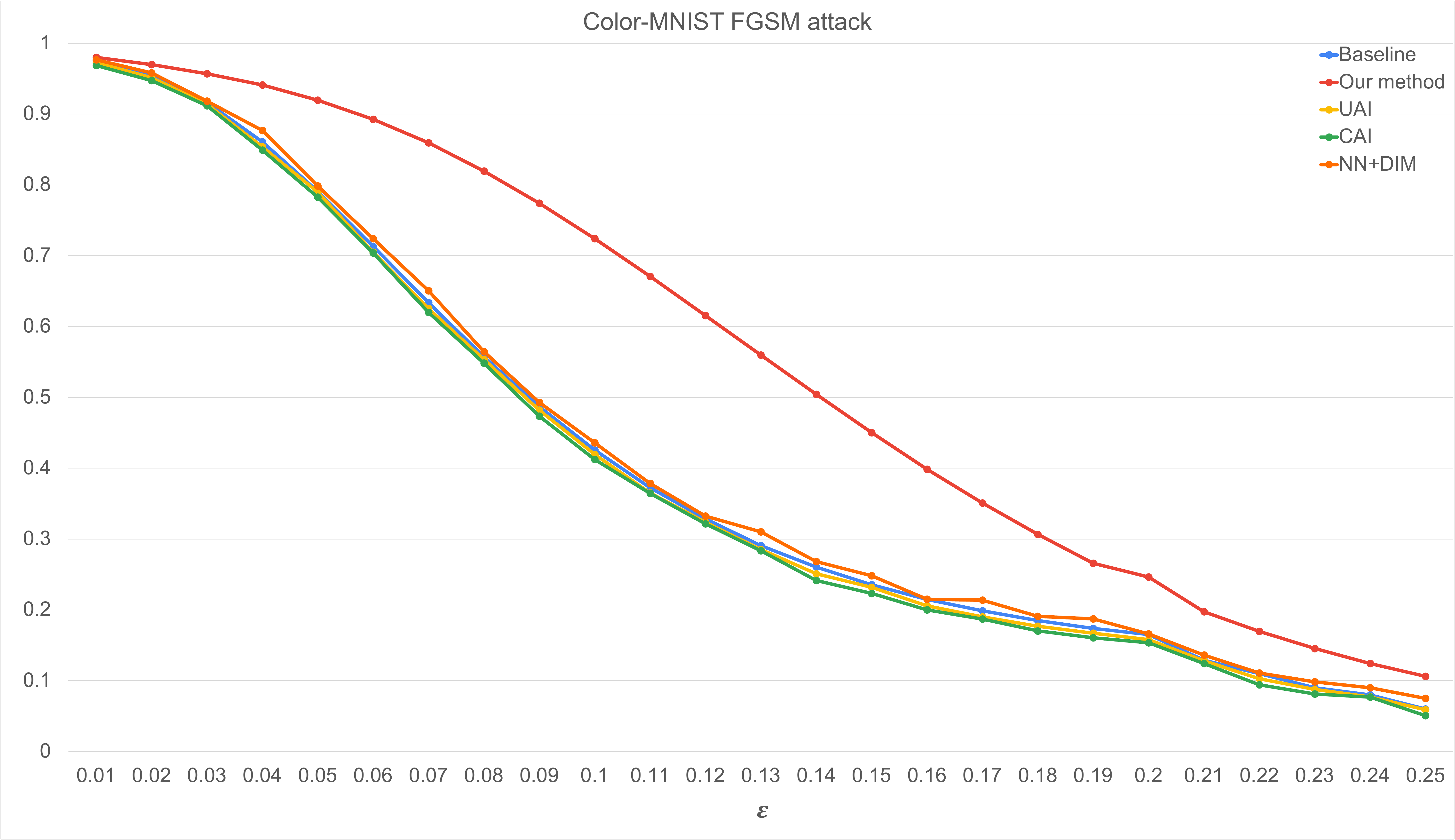}
  \caption{ColorMNIST FGSM attack}
  \label{fig:1}
\end{subfigure}\hfil %
\begin{subfigure}{0.33\textwidth}
  \includegraphics[width=\linewidth,height=0.6\textwidth]{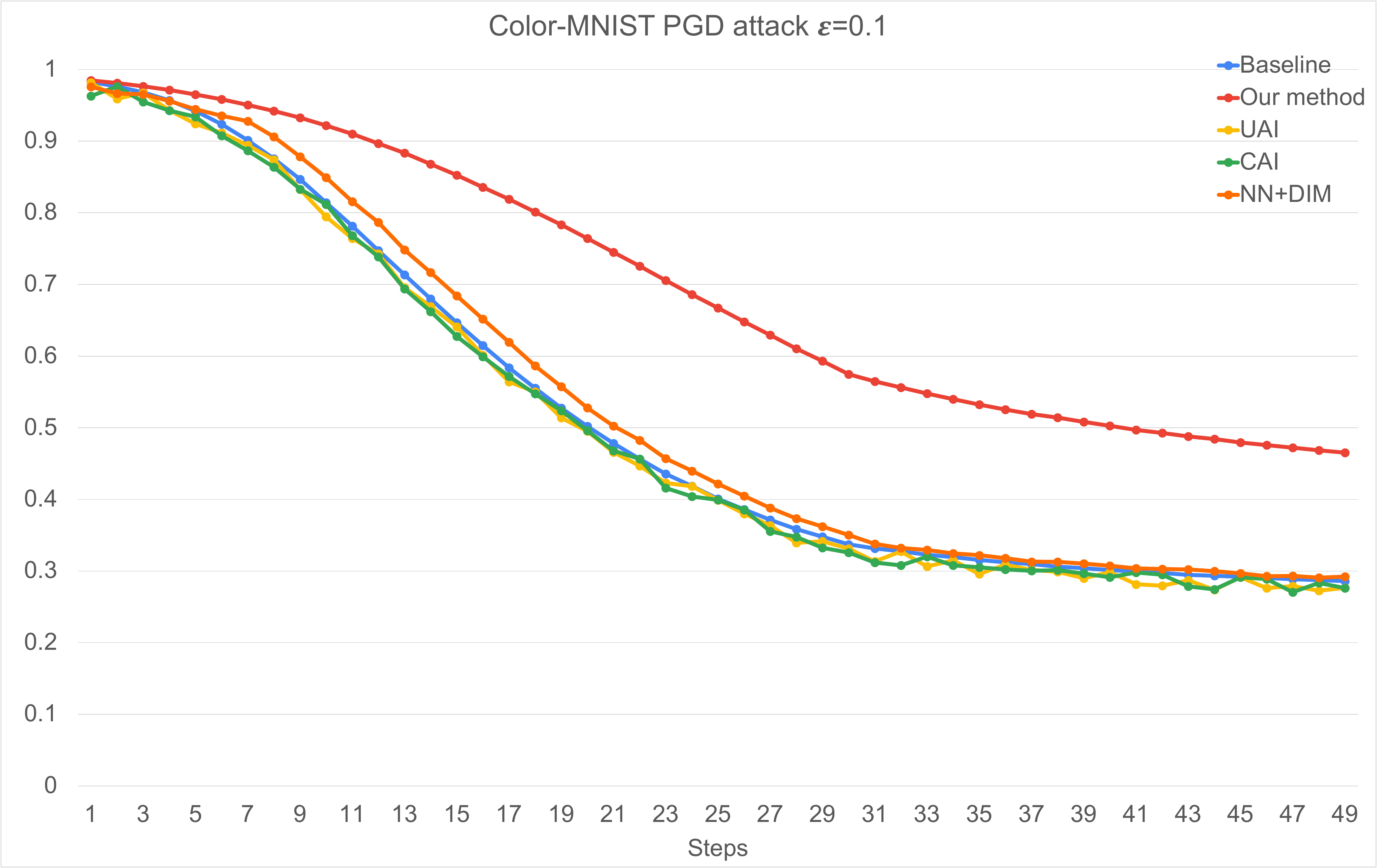}
  \caption{ColorMNIST PGD attack $\epsilon=0.1$}
  \label{fig:2}
\end{subfigure}\hfil %
\begin{subfigure}{0.33\textwidth}
  \includegraphics[width=\linewidth,height=0.6\textwidth]{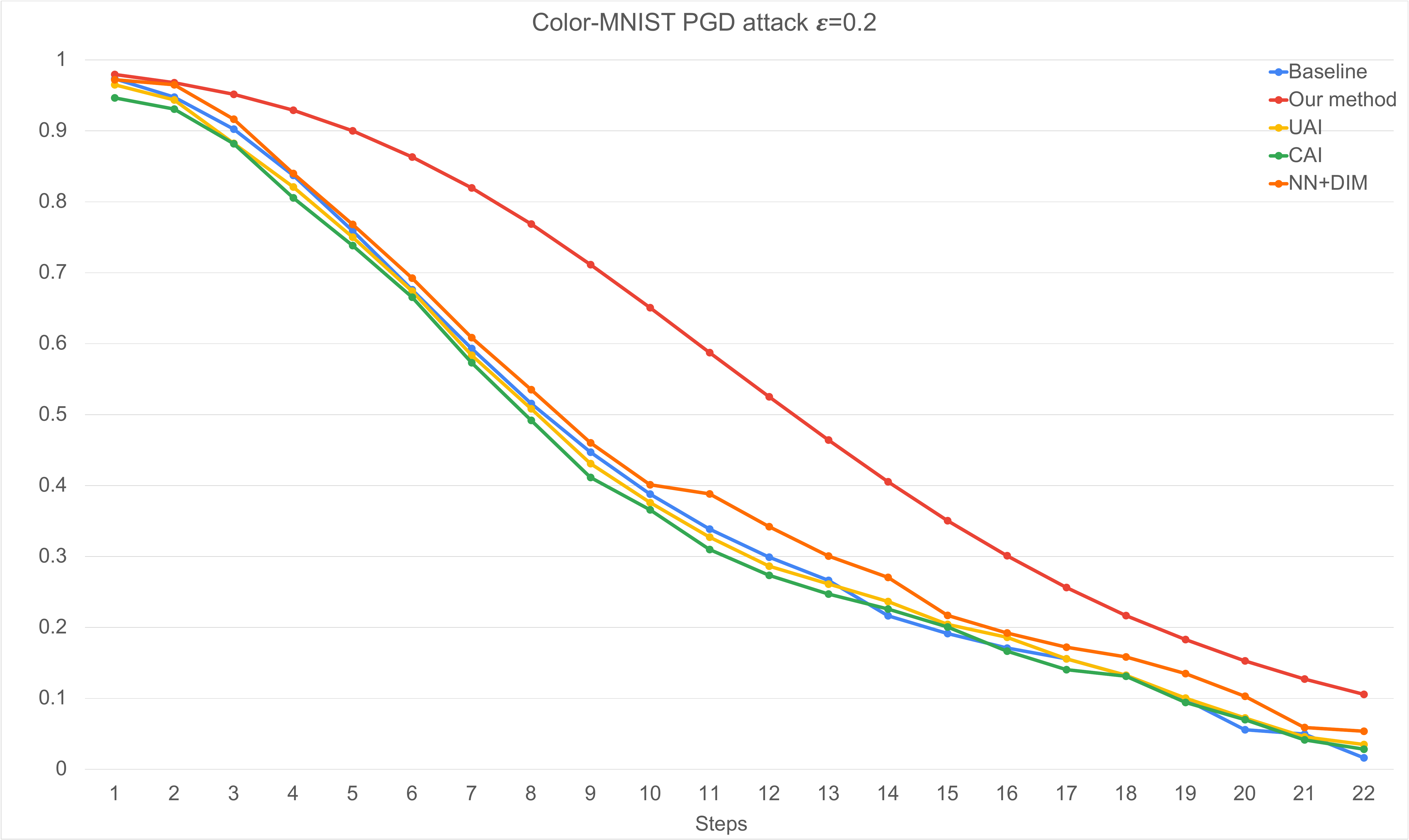}
  \caption{ColorMNIST PGD attack $\epsilon=0.2$}
  \label{fig:3}
\end{subfigure}

\medskip
\begin{subfigure}{0.33\textwidth}
  \includegraphics[width=\linewidth,height=0.6\textwidth]{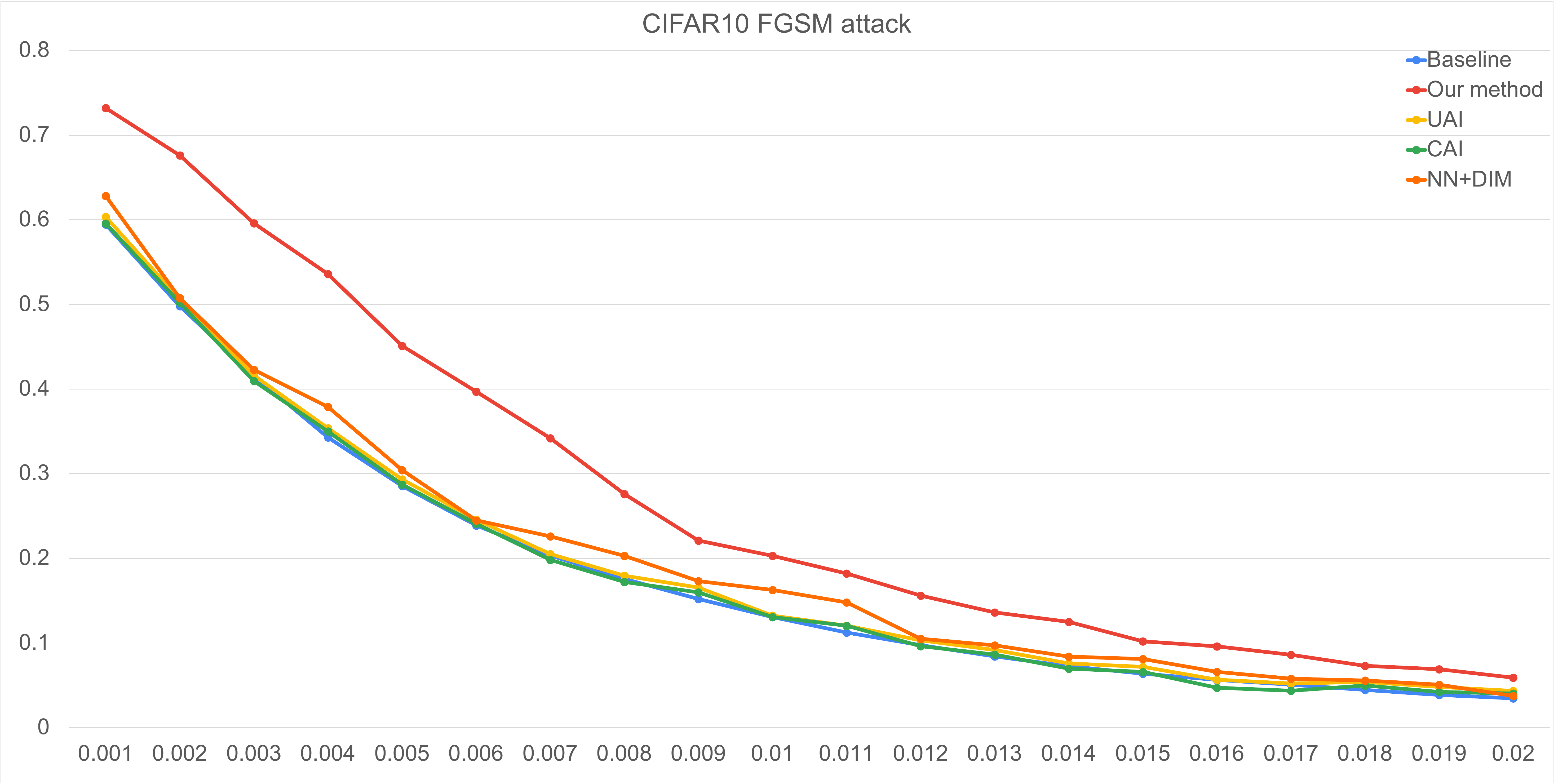}
  \caption{CIFAR10 FGSM attack}
  \label{fig:4}
\end{subfigure}\hfil %
\begin{subfigure}{0.33\textwidth}
  \includegraphics[width=\linewidth,height=0.6\textwidth]{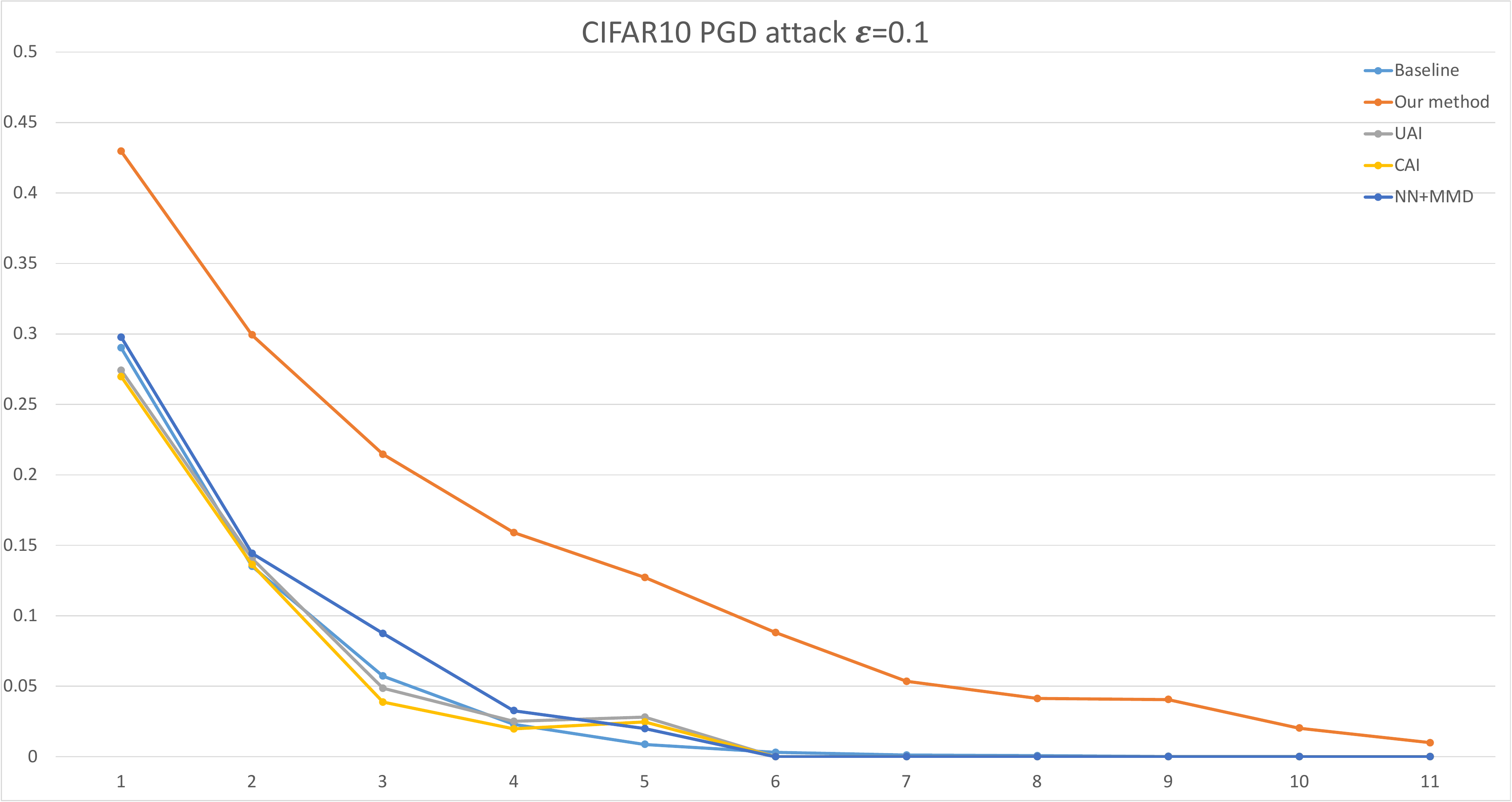}
  \caption{CIFAR10 PGD attack $\epsilon=0.1$}
  \label{fig:5}
\end{subfigure}\hfil %
\begin{subfigure}{0.33\textwidth}
  \includegraphics[width=\linewidth,height=0.6\textwidth]{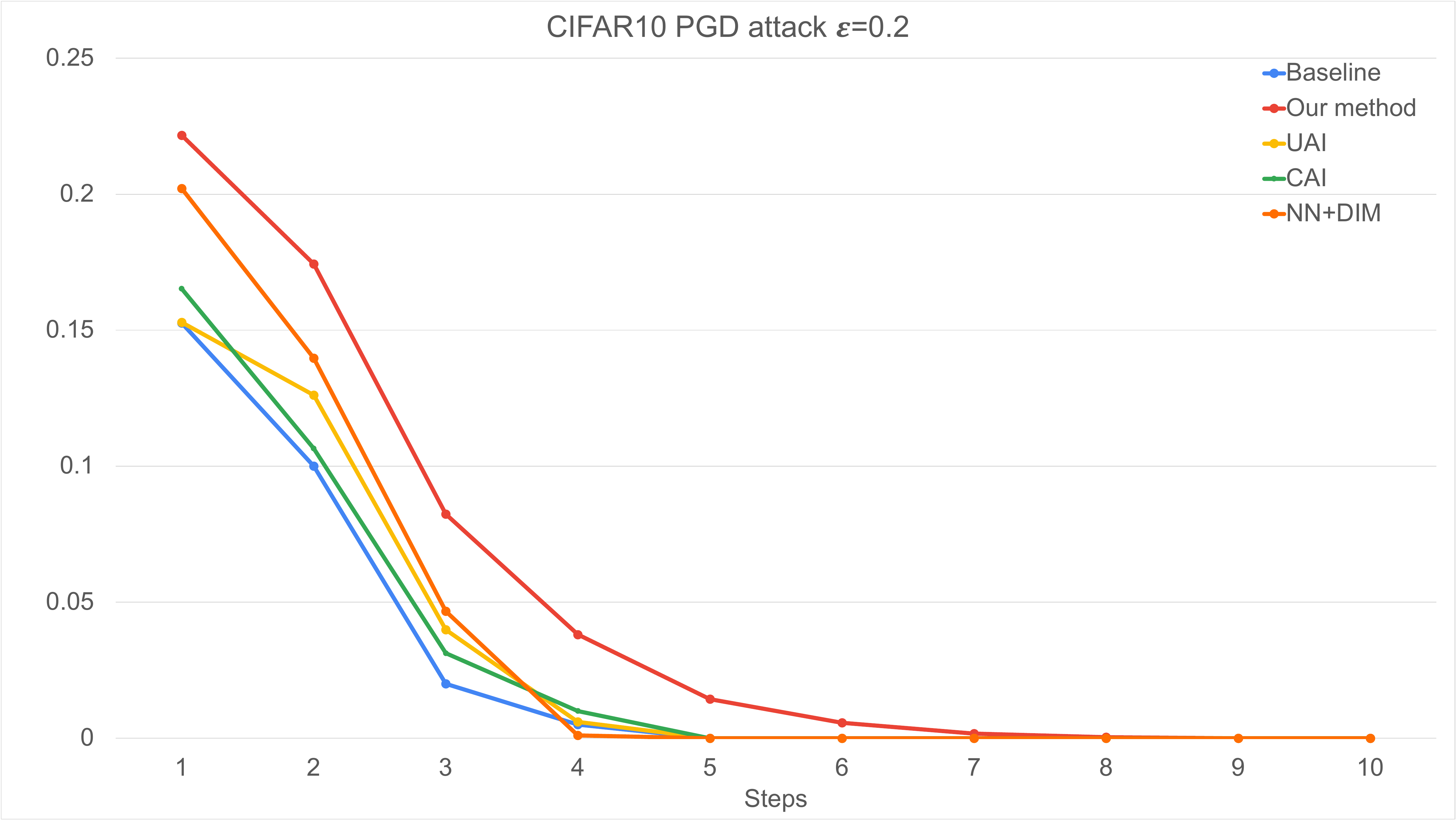}
  \caption{CIFAR10 PGD attack $\epsilon=0.2$}
  \label{fig:6}
\end{subfigure}

\medskip
\begin{subfigure}{0.33\textwidth}
  \includegraphics[width=\linewidth,height=0.6\textwidth]{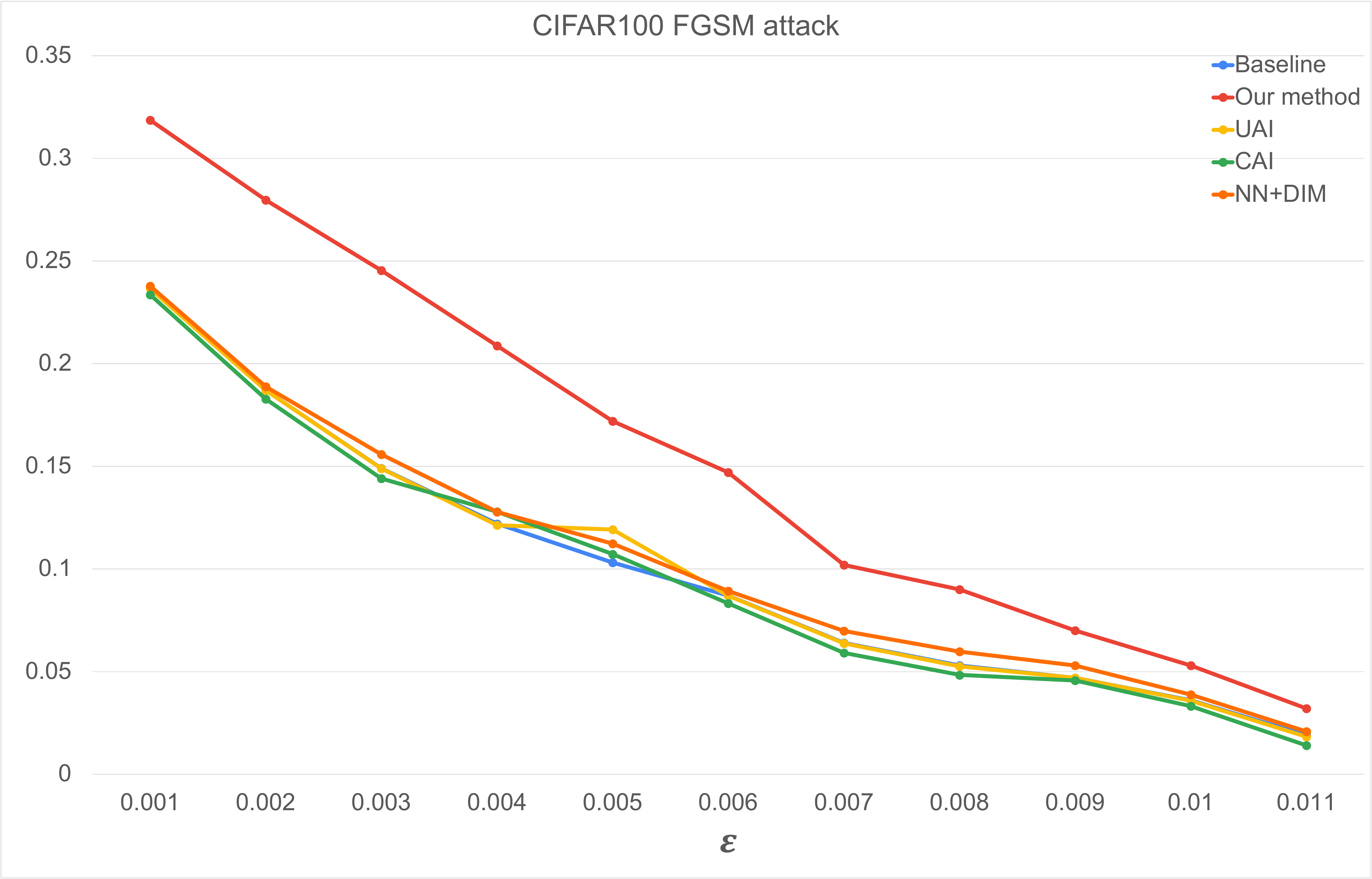}
  \caption{CIFAR100 FGSM attack}
  \label{fig:7}
\end{subfigure}\hfil %
\begin{subfigure}{0.33\textwidth}
  \includegraphics[width=\linewidth,height=0.6\textwidth]{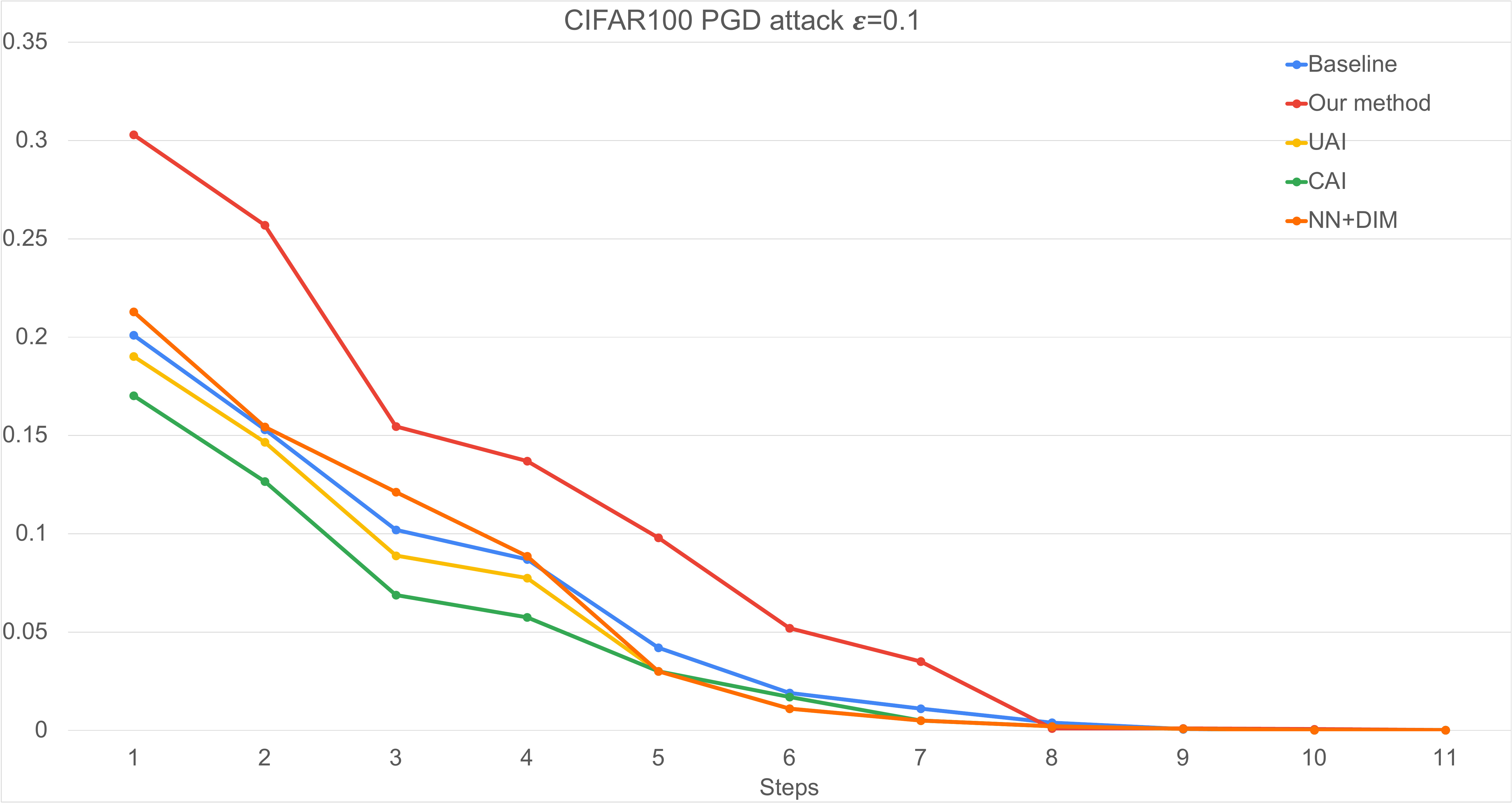}
  \caption{CIFAR100 PGD attack $\epsilon=0.1$}
  \label{fig:8}
\end{subfigure}\hfil %
\begin{subfigure}{0.33\textwidth}
  \includegraphics[width=\linewidth,height=0.6\textwidth]{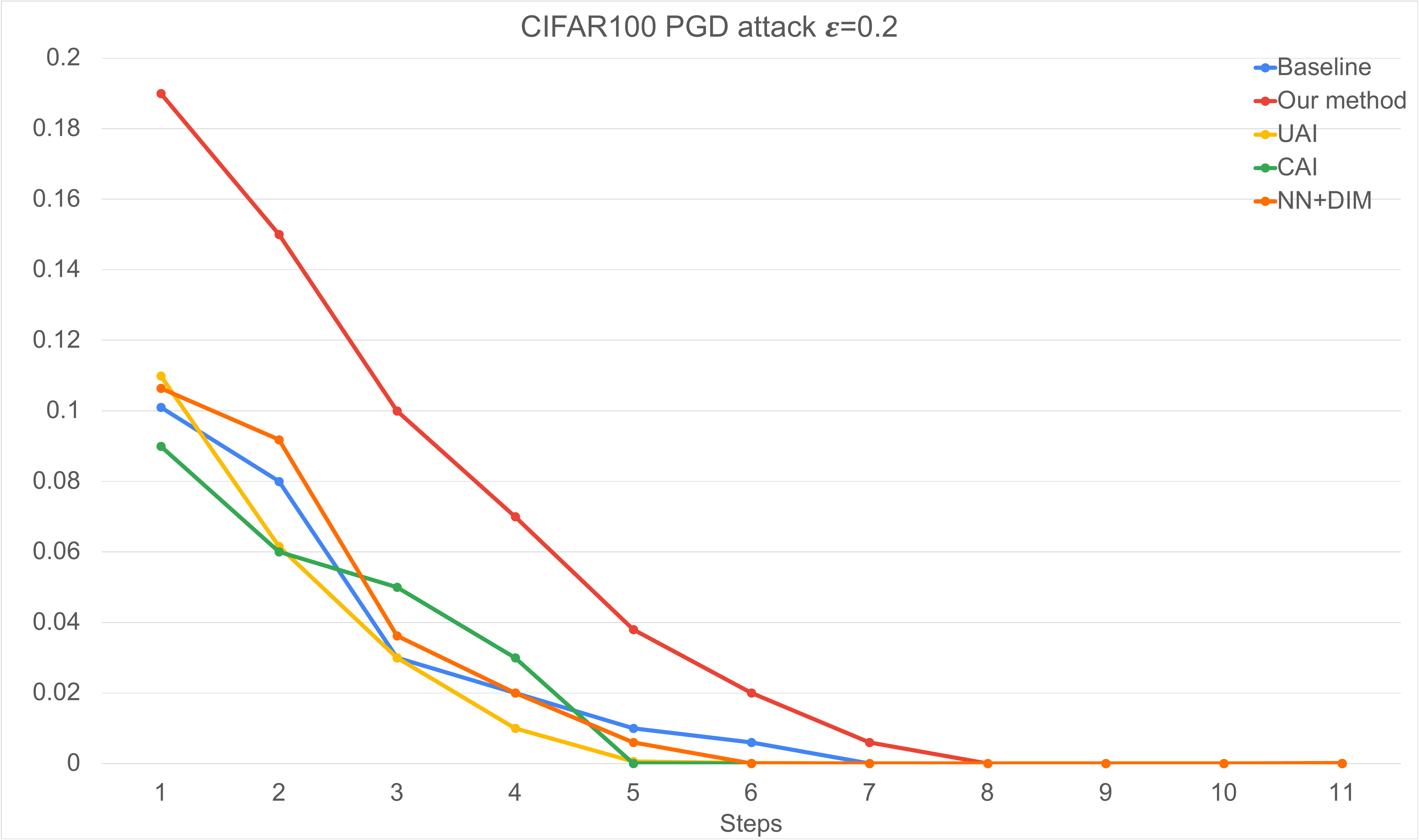}
  \caption{CIFAR100 PGD attack $\epsilon=0.2$}
  \label{fig:9}
\end{subfigure}

\caption{FGSM attacks and PGD attacks results}
\label{fig:adver-attack}
\end{figure}

To test the ability to defend adversarial attack without adversarial augmentation, we first train all models on \textbf{Colored-MNIST}, \textbf{CIFAR10} and \textbf{CIFAR100}. Then, we apply different adversarial attacks on those datasets. The adversarial attack types are:  Fast Gradient Sign Method (FGSM) attack \cite{bib:fgsm}, Projected Gradient Descent (PGD) attack \cite{bib:pgd}, and Carlini \& Wagner (C\&W) attack \cite{bib:cw}. FSGM and PGD attacks results are shown in \Cref{fig:adver-attack} and C\&W attack results are included Appendix.

\begin{table}[]
\centering
\caption{Disentanglement metrics of with different training strategies applied to 3dShapes}
\small
\renewcommand{\arraystretch}{1.3}
{
\begin{adjustbox}{width=0.8\textwidth}
\begin{tabular}{c c| c c c c c}
\hlineB{2}
\textbf{warmup by amount} & \textbf{warmup by difficulty} & \textbf{MIG}     & \textbf{SAP}    & \textbf{IRS}    & \textbf{FVAE} & \textbf{DCI}   \\ \hline
& & 0.492 & 0.096  & 0.661  & 0.902   & 0.697 \\ 
\checkmark &  & 0.512 & 0.126 & 0.674 & 0.944 & 0.781 \\ 
\checkmark & \checkmark & \textbf{0.716} & \textbf{0.156} & \textbf{0.784} & \textbf{0.996}  & \textbf{0.919} \\ \hlineB{2}
\end{tabular}
\end{adjustbox}
\label{table:strategy_algorithm1}
}
\label{table:abla_disen}
\end{table}

\begin{table}[]
\centering
\caption{Performance of different scheme on Colored-MNIST and Rotation-Colored-MNIST}
\label{table:abla_invar}
\renewcommand{\arraystretch}{1.3}
{
\begin{adjustbox}{width=0.8\textwidth}
\begin{tabular}{c|cc|cc}
\hlineB{2}
\multirow{2}{*}{\textbf{Training Scheme}} & \multicolumn{2}{c|}{\textbf{Colored-MNIST}} & \multicolumn{2}{c}{\textbf{Rotation-Colored-MNIST  (65)}} \\
                                          & \textbf{avg acc}   & \textbf{worst acc}   & \textbf{avg acc}           & \textbf{worst acc}          \\ \hline
\textbf{$L_{CE}$}                         & 0.932              & 0.680                & 0.821                      & 0.653                       \\
\textbf{$L_{CE} + L_{contrastive}$}       & 0.935              & 0.732                & 0.842                      & 0.678                       \\
\textbf{$L_{CE} \xleftrightarrow{} L_{contrastive}$}       & \textbf{0.980}     & \textbf{0.904}       & \textbf{0.873}             & \textbf{0.823}              \\ \hlineB{2}
\end{tabular}
\end{adjustbox}
}
\end{table}

\subsection{Ablation Study}
\label{sec:ablation}

\paragraph{Effectiveness of training strategies in disentanglement learning:} 
To prove the effectiveness of the training strategies illustrated in Figure 4, we compare results of three situations: (1) none of those strategies is used, (2) only \emph{warmup by amount} strategy is used, and (3) both strategies are used. As shown in \Cref{table:abla_disen}, using both training strategies clearly outperforms the others. 

\paragraph{Separately training the classifier and the rest of the model:}
To prove the importance of the two-step training as mentioned in Section 4.1, we compare the results of training the entire model together versus separately training classifier and other parts. Further, we also record the results of our model which does not use contrastive loss for feature-level alignment. 
As shown in \Cref{table:abla_invar}, 
either only using CE loss ($L_{CE}$) or training the whole together ($L_{CE}+L_{contrastive}$) will harm the performance of the model. By separately training the classifier and other parts ($L_{CE} \xleftrightarrow{} L_{contrastive} $), the framework has the best results for both average and worst accuracy.

\section{Conclusion}
In this work, we extend the ideas of representation disentanglement  and  representation invariance by combining them to achieve both goals at the same time. By introducing contrastive loss and new invariant regularization loss, we make predictive factor $z_p$ to be more invariant to nuisance and increase both average and worst accuracy on invariant learning tasks. Furthermore, we demonstrate that simultaneously achieving invariant and disentangled representation can increase the performance of adversarial defense comparing to merely using invariant representation learning.

\noindent\textbf{Acknowledgement:} This research is based upon work supported by the Defense Advanced Research Projects Agency (DARPA), under cooperative agreement number \\  HR00112020009. The views and conclusions contained herein should not be interpreted as necessarily representing the official policies or endorsements, either expressed or implied, of DARPA or the U.S. Government. The U.S. Government is authorized to reproduce and distribute reprints for governmental purposes notwithstanding any copyright notation thereon.

\bibliographystyle{splncs04}
\bibliography{reference}

\begin{thebibliography}{10}
\providecommand{\url}[1]{\texttt{#1}}
\providecommand{\urlprefix}{URL }
\providecommand{\doi}[1]{https://doi.org/#1}

\bibitem{bengio2014representation}
Bengio, Y., Courville, A., Vincent, P.: Representation learning: A review and
  new perspectives (2014)

\bibitem{3dshapes18}
Burgess, C., Kim, H.: 3d shapes dataset.
  https://github.com/deepmind/3dshapes-dataset/ (2018)

\bibitem{burgess2018understanding}
Burgess, C.P., Higgins, I., Pal, A., Matthey, L., Watters, N., Desjardins, G.,
  Lerchner, A.: Understanding disentangling in $\beta$-vae (2018)

\bibitem{bib:cw}
Carlini, N., Wagner, D.: Towards evaluating the robustness of neural networks.
  In: 2017 IEEE Symposium on Security and Privacy (SP) (2017)

\bibitem{DBLP:conf/aaai/Castro20}
Castro, P.S.: Scalable methods for computing state similarity in deterministic
  markov decision processes. In: {AAAI} (2020)

\bibitem{DBLP:conf/cvpr/ChenKI20}
Chen, J., Konrad, J., Ishwar, P.: A cyclically-trained adversarial network for
  invariant representation learning. In: {CVPR} Workshops (2020)

\bibitem{chen2019isolating}
Chen, R.T.Q., Li, X., Grosse, R., Duvenaud, D.: Isolating sources of
  disentanglement in variational autoencoders (2019)

\bibitem{bib:dci}
Eastwood, C., Williams, C.K.I.: A framework for the quantitative evaluation of
  disentangled representations. In: ICLR (2018),
  \url{https://openreview.net/forum?id=By-7dz-AZ}

\bibitem{bib:dann}
Ganin, Y., Ustinova, E., Ajakan, H., Germain, P., Larochelle, H., Laviolette,
  F., Marchand, M., Lempitsky, V.: Domain-adversarial training of neural
  networks. The Journal of Machine Learning Research  (2016)

\bibitem{gondal2019transfer}
Gondal, M.W., Wüthrich, M., Đorđe Miladinović, Locatello, F., Breidt, M.,
  Volchkov, V., Akpo, J., Bachem, O., Schölkopf, B., Bauer, S.: On the
  transfer of inductive bias from simulation to the real world: a new
  disentanglement dataset (2019)

\bibitem{bib:fgsm}
Goodfellow, I.J., Shlens, J., Szegedy, C.: Explaining and harnessing
  adversarial examples (2014), \url{http://arxiv.org/abs/1412.6572}, cite
  arxiv:1412.6572

\bibitem{he2015deep}
He, K., Zhang, X., Ren, S., Sun, J.: Deep residual learning for image
  recognition (2015)

\bibitem{bib:betaVAE}
Higgins, I., Matthey, L., Pal, A., Burgess, C., Glorot, X., Botvinick, M.,
  Mohamed, S., Lerchner, A.: beta-vae: Learning basic visual concepts with a
  constrained variational framework. In: {ICLR} (2017)

\bibitem{bib:uai}
Jaiswal, A., Wu, R.Y., Abd-Almageed, W., Natarajan, P.: Unsupervised
  adversarial invariance. In: Advances in Neural Information Processing Systems
  31 (2018)

\bibitem{bib:super_contras}
Khosla, P., Teterwak, P., Wang, C., Sarna, A., Tian, Y., Isola, P., Maschinot,
  A., Liu, C., Krishnan, D.: Supervised contrastive learning. CoRR
  \textbf{abs/2004.11362} (2020), \url{https://arxiv.org/abs/2004.11362}

\bibitem{bib:factorvae}
Kim, H., Mnih, A.: Disentangling by factorising. In: ICML (2018),
  \url{http://proceedings.mlr.press/v80/kim18b.html}

\bibitem{bib:vae}
Kingma, D.P., Welling, M.: Auto-encoding variational bayes. In: {ICLR} (2014)

\bibitem{bib:dipvae}
Kumar, A., Sattigeri, P., Balakrishnan, A.: Variational inference of
  disentangled latent concepts from unlabeled observations. ICLR  (2018)

\bibitem{bib:pgd}
Kurakin, A., Goodfellow, I.J., Bengio, S.: Adversarial machine learning at
  scale. CoRR  \textbf{abs/1611.01236} (2016),
  \url{http://arxiv.org/abs/1611.01236}

\bibitem{lecun-mnisthandwrittendigit-2010}
LeCun, Y., Cortes, C.: {MNIST} handwritten digit database  (2010),
  \url{http://yann.lecun.com/exdb/mnist/}

\bibitem{bib:nnmmd}
Li, Y., Swersky, K., Zemel, R.: Learning unbiased features. arXiv preprint
  arXiv:1412.5244  (2014)

\bibitem{locatello2019challenging}
Locatello, F., Bauer, S., Lucic, M., Rätsch, G., Gelly, S., Schölkopf, B.,
  Bachem, O.: Challenging common assumptions in the unsupervised learning of
  disentangled representations (2019)

\bibitem{bib:adavae}
Locatello, F., Poole, B., Raetsch, G., Sch{\"o}lkopf, B., Bachem, O.,
  Tschannen, M.: Weakly-supervised disentanglement without compromises. In:
  ICML (2020), \url{http://proceedings.mlr.press/v119/locatello20a.html}

\bibitem{bib:semi-disen}
Locatello, F., Tschannen, M., Bauer, S., Rätsch, G., Schölkopf, B., Bachem,
  O.: Disentangling factors of variations using few labels. In: ICLR (2020),
  \url{https://openreview.net/forum?id=SygagpEKwB}

\bibitem{bib:vfae}
Louizos, C., Swersky, K., Li, Y., Welling, M., Zeme, R.: The variational fair
  autoencoder. In: ICLR (2016)

\bibitem{bib:cvib}
Moyer, D., Gao, S., Brekelmans, R., Galstyan, A., Ver~Steeg, G.: Invariant
  representations without adversarial training. In: Advances in Neural
  Information Processing Systems 31 (2018)

\bibitem{Sagawa*2020Distributionally}
Sagawa*, S., Koh*, P.W., Hashimoto, T.B., Liang, P.: Distributionally robust
  neural networks. In: ICLR (2020),
  \url{https://openreview.net/forum?id=ryxGuJrFvS}

\bibitem{bib:irmi}
Sanchez, E., Serrurier, M., Ortner, M.: Learning disentangled representations
  via mutual information estimation. In: ECCV (2020)

\bibitem{DBLP:journals/corr/abs-1905-12506}
van Steenkiste, S., Locatello, F., Schmidhuber, J., Bachem, O.: Are
  disentangled representations helpful for abstract visual reasoning? CoRR
  \textbf{abs/1905.12506} (2019), \url{http://arxiv.org/abs/1905.12506}

\bibitem{suter2019robustly}
Suter, R., Đorđe Miladinović, Schölkopf, B., Bauer, S.: Robustly
  disentangled causal mechanisms: Validating deep representations for
  interventional robustness (2019)

\bibitem{Wang_2021_CVPR}
Wang, F., Liu, H.: Understanding the behaviour of contrastive loss. In: CVPR
  (2021)

\bibitem{bib:cai}
Xie, Q., Dai, Z., Du, Y., Hovy, E., Neubig, G.: Controllable invariance through
  adversarial feature learning. In: Advances in Neural Information Processing
  Systems 30 (2017)

\end{thebibliography}
\clearpage
\section{Appendix}
\counterwithin{figure}{section}
\counterwithin{table}{section}
\setcounter{figure}{0}
\setcounter{table}{0}
\setcounter{page}{1}
\subsection{Why reconstruction job is necessary for independent predictive factors}
\label{sec:proof}
This proof is largely dependent on proof used in \cite{locatello2019challenging} and the proof is shown below:

By the assumption, we hope the latent factors can be separated into two part $z_p$ and $z_n$. $z_p$ and $z_n$ are expected to be independent to each other. We have :
$$ p(z) = p(z_p) \cdot p(z_n) $$

Thus, if we choose any latent factor $z_i$ from $z_p$, it should be independent to any other latent factor $z_j$ chosen from $z_n$. $z_i \indep z_j$.

It can be claimed that there exists an infinite family of bijective functions $f: supp(z)
\xrightarrow{}supp(z)$ such that $ \frac{\partial f_i(u)}{\partial u_j} \neq  0$ almost everywhere for all $i$ in $z_p$ and $j$ from $z_n$ (i.e., $z$ and $f(z)$ are completely entangled) and $P(z \leq u) =P(f(z) \leq u)$ for all $u \in supp(z)$ (i.e., they have the same marginal distribution). Since the unsupervised method only has access to observations $x$ and $y$, it hence cannot distinguish between the two equivalent generative models and thus has to be entangled to at least one of them.

We first choose two bijective functions $g_i(v_i)= P(z_i \leq v_i)$  and $g_j(v_j)= P(z_j \leq v_j)$. By construction $g(z)=g(z_i) \cdot g(z_j)$ is a 2-dimensional uniform distribution. Similarly, consider function $h_i(v_i) = \psi^{-1}(v_i)$ and $h_j(v_j) = \psi^{-1}(v_j)$. Where,

$\psi(\cdot)$ is the cumulative density function of standard normal distribution. By this further construction, the random variable $h(g(z))$ is a 2-dimensional standard normal distribution.

Let $A \in R^{2 \times 2}$ be an arbitrary orthogonal matrix with $A_{km} \neq 0$ for all $k=1,2$ and $m=1,2$. An infinite family of such matrices can be constructed using a Householder transformation: Choose an arbitrary $\alpha \in (0,0.5)$and consider the vector $v$ with $v_1=\sqrt{\alpha}$ and $v_2=\sqrt{1-\alpha}$. By construction, we have $\mathbf{v}^{T}\mathbf{v}=1$. Define the matrix $A=\mathbf{I}_{2} - 2 \mathbf{v}\mathbf{v}^T$. Furthermore, $A$ is orthogonal since

$$A^{T}A=(\mathbf{I}_2 - 2\mathbf{v}\mathbf{v}^T)^{T}(\mathbf{I}_2 - 2\mathbf{v}\mathbf{v}^T)=\mathbf{I}_2 - 4\mathbf{v}\mathbf{v}^{T} + 4\mathbf{v}(\mathbf{v}^T\mathbf{v})\mathbf{v}^T = \mathbf{I}_2$$.

Since $A$ is orthogonal,  it is invertible and thus defines a bijective linear operator. The random variable $Ah(g(z)) \in R^2$ is hence an independent, multivariate standard normal distribution since the covariance matrix $A^T A$ is equal to $I_2$.

Since $h$ is bijective, it follows that $h^{-1}(Ah(g(z)))$is an independent 2-dimensional uniform distribution.  Define the function $f: supp(z) \xrightarrow{}supp(z)$ 
$$f(u) =g^{-1}(h^{-1}(Ah(g(u))))$$

and note that by definition $f(z)$ has the same marginal distribution as $z$ under $P$,i.e.,$P(z \leq u) =P(f(z) \leq u)$for all $u$ .Finally, for almost every $u \in supp(z)$, it holds that

$$\frac{\partial f_i(u)}{\partial u_j} \neq 0, $$
Since $A$ was chosen arbitrarily, there exists an infinite family of such function $f$. 

To overcomes this problem, we need to do similar reconstruction job like we do for disentangled representation learning, where introducing supervision signal to enforce independence between $z_p$ and $z_n$ is necessary. However, previous works \cite{bib:uai,bib:cai,bib:vfae,bib:cvib} fail to do that. The detail of the process is described in Section 3.3.

\subsection{Other Adversarial attack results}
 In this section, We record the Carlini \& Wangner (C\&W) attack results with varing inital constant $c$ in \Cref{table:adver1}.

\begin{table}[H]
\caption{Carlini \& Wangner (C\&W) attack results}
\label{table:adver1}
\centering
\renewcommand{\arraystretch}{1.3}
{
\begin{adjustbox}{width=0.8\textwidth}
\begin{tabular}{l|ccl|cc|c}
\hline
\multirow{2}{*}{Model} & \multicolumn{3}{c|}{Color-MNIST} & \multicolumn{2}{c|}{CIFR10} & CIFAR100 \\
                       & c = 0.01   & c = 0.1   & c = 1   & c = 0.01      & c = 0.1     & c = 0.01 \\ \cline{2-7} 
Baseline               & 0.923      & 0.616     & 0.282   & 0.301         & 0.112       & 0.103    \\
UAI~\cite{bib:uai}                  & 0.916      & 0.607     & 0.279   & 0.310         & 0.083       & 0.097    \\
CAI~\cite{bib:cai}                   & 0.902      & 0.594     & 0.267   & 0.319         & 0.105       & 0.081    \\
NN+DIM~\cite{bib:irmi}               & 0.917      & 0.638     & 0.295   & 0.296         & 0.135       & 0.121    \\
Our method             & 0.945      & 0.792     & 0.496   & 0.462         & 0.204       & 0.213    \\ \hline
\end{tabular}
\end{adjustbox}
}

\end{table}

\subsection{Visualization of latent space}
\begin{figure}[!htp]
    \centering
     \begin{subfigure}[b]{0.49\textwidth}
     \centering
         \includegraphics[width=1\textwidth]{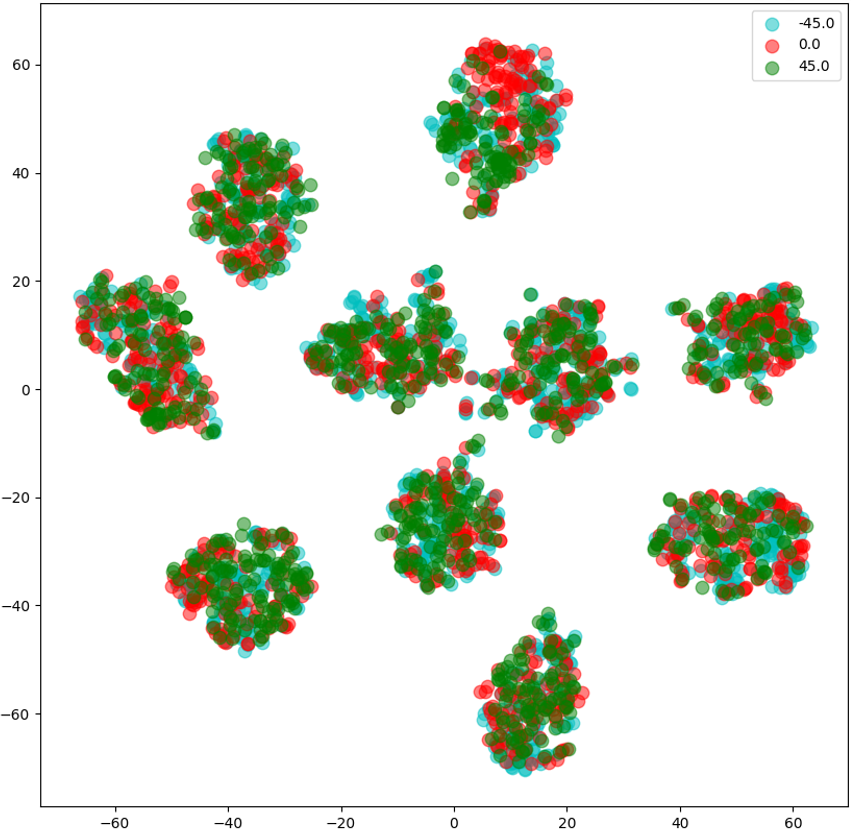}
         \caption{$z_p$ embedding}
     \end{subfigure}
     \hfill
     \begin{subfigure}[b]{0.49\textwidth}
     \centering
         \includegraphics[width=1\textwidth]{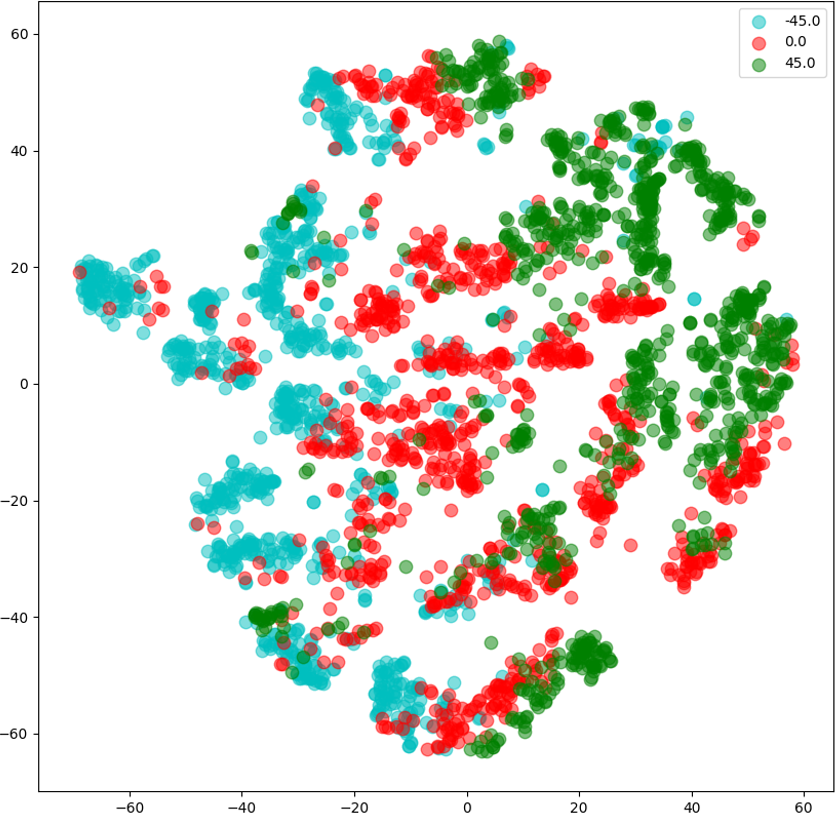}
         \caption{$z_{nu}$ embedding}
    \end{subfigure}
  \caption{t-SNE visualization of $z_p$ and $z_{nu}$ embeddings of Color-Rotation-MNIST images colored by rotation angle. As desired, the $z_p$ embedding does not encode rotation information, which migrates to $z_{nu}$.}
     \label{fig:tsne}
\end{figure}

To illustrate the performance of the results of the model, we visualize the latent representation by different methods.
We first visualize the t-SNE results of $z_p$ and $z_{nu}$ which we tested on Color-Rotation-MNIST in \Cref{fig:tsne}.

\begin{figure}[!htp]
    \centering
     \begin{subfigure}[b]{0.49\textwidth}
     \centering
         \includegraphics[width=1\textwidth]{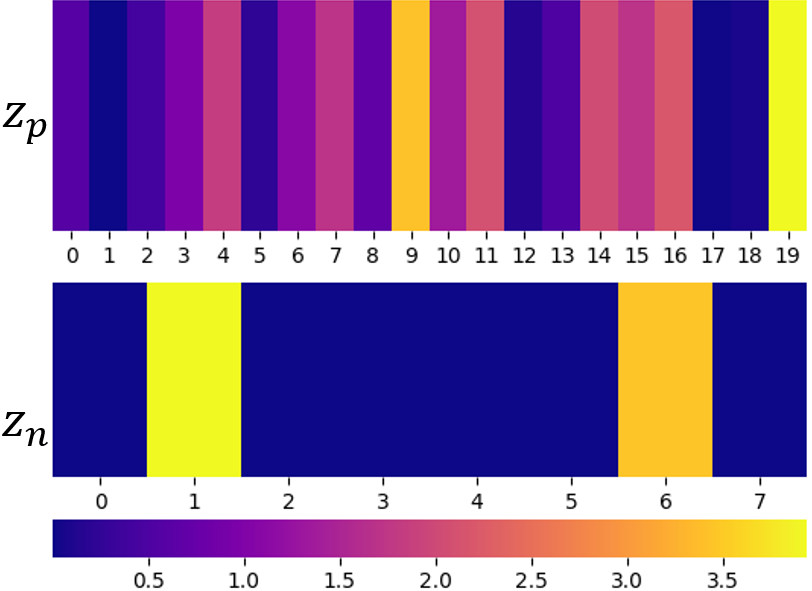}
         \caption{latent factors change by changing the background color and digit color}
     \end{subfigure}
     \hfill
     \begin{subfigure}[b]{0.49\textwidth}
     \centering
         \includegraphics[width=1\textwidth]{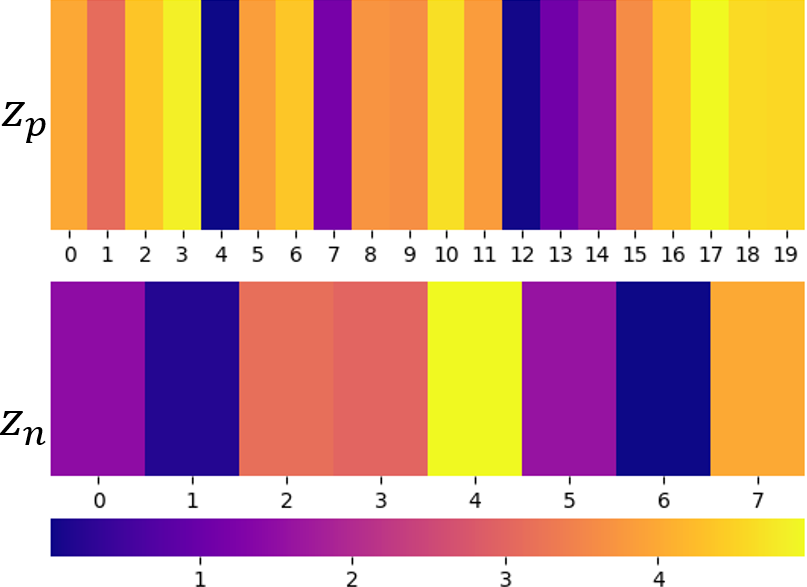}
         \caption{latent factors change by changing the digit number and make the color unchanged}
    \end{subfigure}
  \caption{Heat map visualization for Colored-MNIST dataset}
     \label{fig:heatmap1}
\end{figure}

\begin{figure}[!htp]
    \centering
     \begin{subfigure}[b]{0.32\textwidth}
     \centering
         \includegraphics[width=1\textwidth]{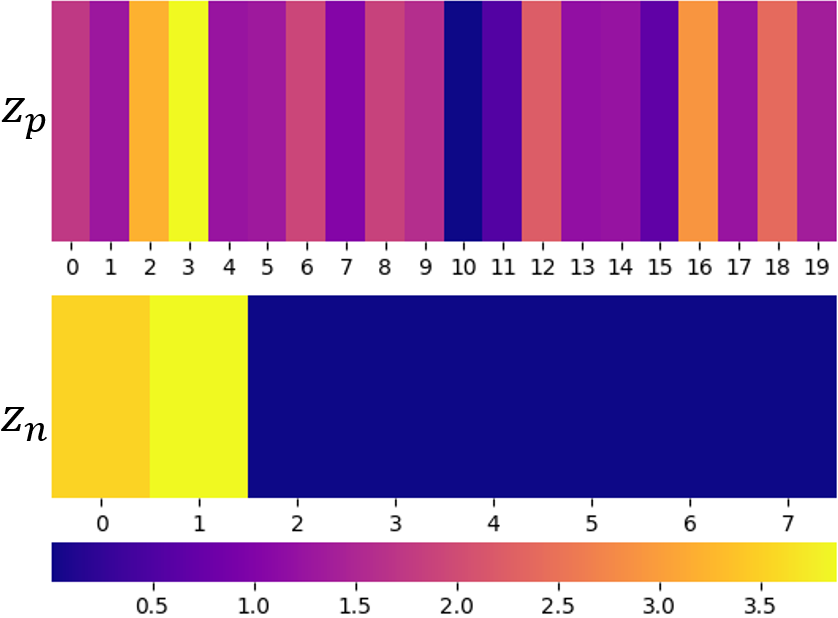}
         \caption{latent factors change by changing the color}
     \end{subfigure}
     \hfill
     \begin{subfigure}[b]{0.32\textwidth}
     \centering
         \includegraphics[width=1\textwidth]{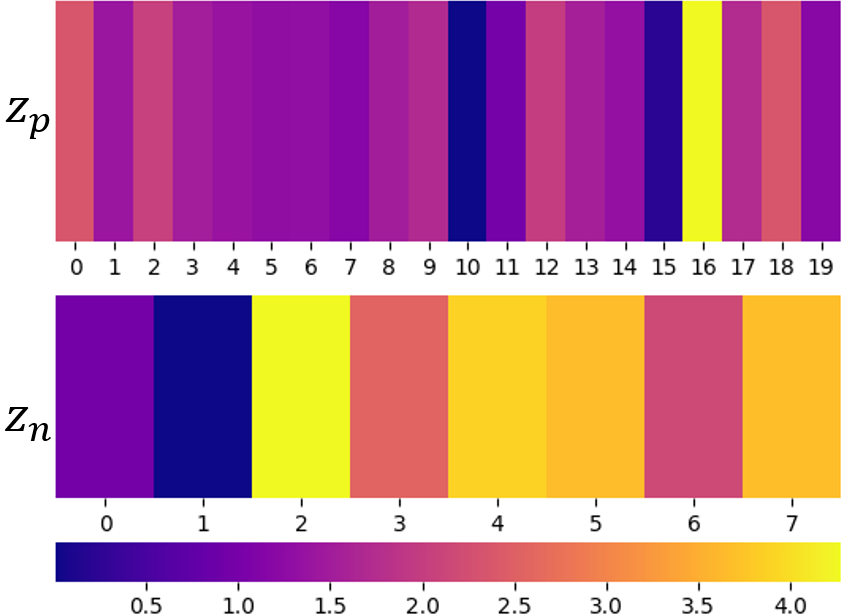}
         \caption{latent factors change by changing the angle}
    \end{subfigure}
    \hfill
     \begin{subfigure}[b]{0.32\textwidth}
     \centering
         \includegraphics[width=1\textwidth]{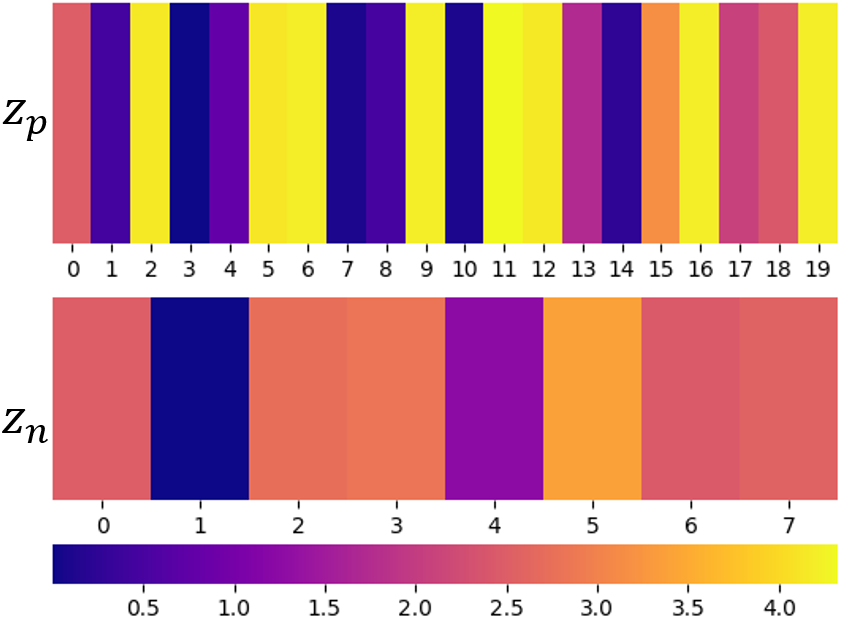}
         \caption{latent factors change by changing the angle}
    \end{subfigure}
    
  \caption{Heat map visualization for Rotation-Colored-MNIST dataset}
  \label{fig:heatmap2}
\end{figure}

Further, we visualize the heatmap of latent space change by giving the model with different inputs. In \Cref{fig:heatmap1}, we visualize the latent factors change of model tested on Colored-MNIST and visualize the latent factors change of model tested on Rotation-Colored-MNIST in \Cref{fig:heatmap2}.

\begin{figure}[!htp]
    \centering
    \includegraphics[width=1\linewidth]{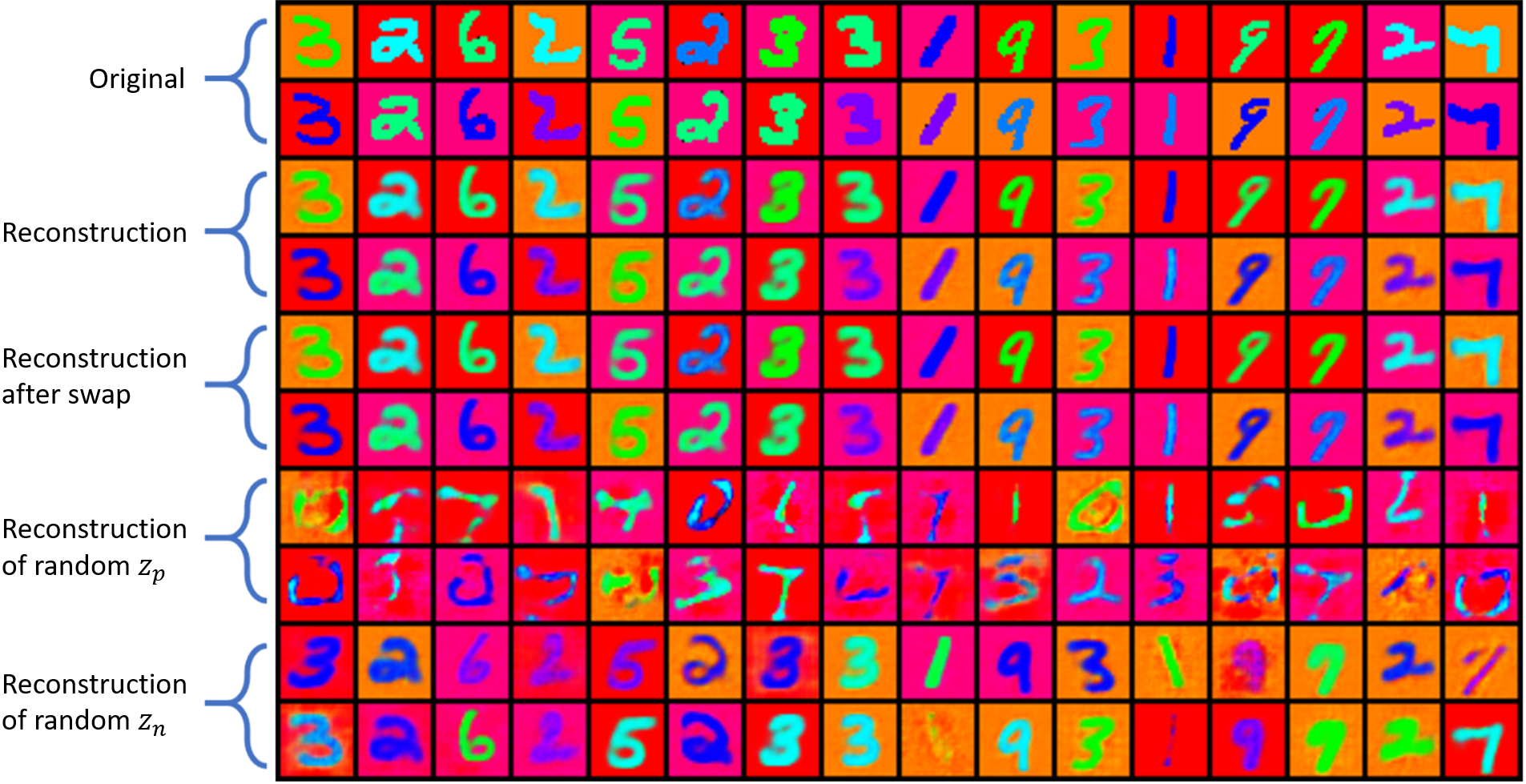}
    \caption{Reconstruction results of Colored-MNIST data}
    \label{fig:reconstruction}
\end{figure}

We finally visualize the results of reconstruction of model we tested on Colored-MNIST in \Cref{fig:reconstruction}. Images in line 1-2 are original images used for training. Images in line 3-4 are reconstructions which are expected to be same with original inputs. Images in line 5-6 are reconstructions after swapping operation which are also expected to be same with original inputs and reconstruction in line 2-3. Images in line 7-8 are decoded from $[rand(z_p),z_n]$, where $rand(z_p)$ is normal random noise. Since we random sample $z_p$, the outputs of decoder should only contains the color information and unrecognized digits. In the contrary, images in line 9-10, we randomly sampled $z_n$ and the latent factors for decoding is $[z_p,rand(z_n)]$ . Therefore, images in line 9-10 should have same digit pattern with original inputs but have random digit color and background color. Besides, other nuisance factors like hand-writting style and dilated or eroded shapes will also change since $z_n$ has split $z_{nu}$ which contains unknown nusiance factors information.

\end{document}